\documentclass[preprint,12pt]{elsarticle}

\usepackage{amssymb}
\usepackage{amsmath}
\usepackage{multirow} 
\usepackage{booktabs} 
\usepackage{algorithmic}
\usepackage{graphicx}
\usepackage{algorithm}
\makeatletter
\newcommand{\removelatexerror}{\let\@latex@error\@gobble}
\makeatother
\usepackage{subcaption}

\journal{xxxxxx}

\begin{document}

\begin{frontmatter}

%% Title, authors and addresses

%% use the tnoteref command within \title for footnotes;
%% use the tnotetext command for theassociated footnote;
%% use the fnref command within \author or \affiliation for footnotes;
%% use the fntext command for theassociated footnote;
%% use the corref command within \author for corresponding author footnotes;
%% use the cortext command for theassociated footnote;
%% use the ead command for the email address,
%% and the form \ead[url] for the home page:
%% \title{Title\tnoteref{label1}}
%% \tnotetext[label1]{}
%% \author{Name\corref{cor1}\fnref{label2}}
%% \ead{email address}
%% \ead[url]{home page}
%% \fntext[label2]{}
%% \cortext[cor1]{}
%% \affiliation{organization={},
%%             addressline={},
%%             city={},
%%             postcode={},
%%             state={},
%%             country={}}
%% \fntext[label3]{}

\title{FedGRec: Dynamic Spatio-Temporal Federated Graph Learning for Secure and Efficient Cross-Border Recommendations}

%% use optional labels to link authors explicitly to addresses:
%% \author[label1,label2]{}
%% \affiliation[label1]{organization={},
%%             addressline={},
%%             city={},
%%             postcode={},
%%             state={},
%%             country={}}
%%
%% \affiliation[label2]{organization={},
%%             addressline={},
%%             city={},
%%             postcode={},
%%             state={},
%%             country={}}

\author[1]{Zhizhong Tan}\ead{zhizhongtan@163.com}
%\author[1,2]{Jiexin Zheng\fnref{equal}}\ead{3230002613@student.must.edu.mo}
\author[1,2]{Jiexin Zheng}\ead{3230002613@student.must.edu.mo}
\author[1]{Xingxing Yang}\ead{3220006956@student.must.edu.mo}
%\author[1]{Yan Liu}\ead{2109853jin30001@student.must.edu.mo}
%\author[1]{Mudi Xu}\ead{2109853gin30002@student.must.edu.mo}
%\author[1]{Ningning Zhang}\ead{jojopdq@gmail.com}
\author[1]{Chi Zhang}\ead{2109853PIN30001@student.must.edu.mo}

%\author[1]{Xinning Xiong}\ead{3220004627@student.must.edu.mo}
%\author[1]{Ningning Zhang}\ead{jojopdq@gmail.com}
\author[1]{Weiping Deng}\ead{dengweipingok@126.com}
\author[1]{Wenyong Wang\corref{corresponding}}
\ead{wywang@must.edu.mo}

%% Author affiliation
\affiliation[1]{organization={School of Computer Science and Engineering, Macau University of Science and Technology},
            %addressline={}, 
            city={Macau},
%          citysep={}, % Uncomment if no comma needed between city and postcode
            postcode={999074}, 
            %state={},
            country={China}} 

\affiliation[2]{organization={Guangdong Institute of Intelligence Science and Technology},%Department and Organization
            %addressline={}, 
            city={Zhuhai},
            postcode={519031}, 
            %state={},
            country={China}}
            
\cortext[corresponding]{Corresponding author.}

%% Abstract
\begin{abstract}
Due to the highly sensitive nature of certain data in cross-border sharing, collaborative cross-border recommendations and data sharing are often subject to stringent privacy protection regulations, resulting in insufficient data for model training. Consequently, achieving efficient cross-border business recommendations while ensuring privacy security poses a significant challenge. Although federated learning has demonstrated broad potential in collaborative training without exposing raw data, most existing federated learning-based GNN training methods still rely on federated averaging strategies, which perform suboptimally on highly heterogeneous graph data. To address this issue, we propose FedGRec, a privacy-preserving federated graph learning method for cross-border recommendations. FedGRec captures user preferences from distributed multi-domain data to enhance recommendation performance across all domains without privacy leakage. Specifically, FedGRec leverages collaborative signals from local subgraphs associated with users or items to enrich their representation learning. Additionally, it employs dynamic spatiotemporal modeling to integrate global and local user preferences in real time based on business recommendation states, thereby deriving the final representations of target users and candidate items. By automatically filtering relevant behaviors, FedGRec effectively mitigates noise interference from unreliable neighbors. Furthermore, through a personalized federated aggregation strategy, FedGRec adapts global preferences to heterogeneous domain data, enabling collaborative learning of user preferences across multiple domains. Extensive experiments on three datasets demonstrate that FedGRec consistently outperforms competitive single-domain and cross-domain baselines while effectively preserving data privacy in cross-border recommendations.
\end{abstract}

%%Graphical abstract
% \begin{graphicalabstract}
% %\includegraphics{grabs}
% \end{graphicalabstract}

%%Research highlights
% \begin{highlights}
% \item Research highlight 1
% \item Research highlight 2
% \end{highlights}

%% Keywords
\begin{keyword}
%% keywords here, in the form: keyword \sep keyword

%% PACS codes here, in the form: \PACS code \sep code

%% MSC codes here, in the form: \MSC code \sep code
%% or \MSC[2008] code \sep code (2000 is the default)
Cross-Border \sep Privacy-preserving \sep Spatio-Temporal \sep Federated Recommendation \sep GNN.
\end{keyword}

\end{frontmatter}

\section{Introduction}\label{Introduction}

With the gradual opening of cross-border data and the increasing volume of international business operations, inter-institutional business recommendation systems have gained extensive application scenarios and significant value in cross-border data sharing. By leveraging diverse cross-border datasets, institutions can not only provide customers with more accurate and personalized services—enhancing user experience and satisfaction—but also optimize risk management processes, drive business innovation and product improvement, support compliance and regulatory efforts, and foster collaboration and resource sharing among institutions across different regions.
However, conventional recommendation systems typically require the collection of user and item attributes, along with extensive interaction data, to model user preferences and achieve precise recommendations. Consequently, as users place greater emphasis on data security and privacy protection regulations become increasingly stringent, a critical challenge for cross-border recommendation systems lies in effectively utilizing fragmented data to continuously improve model performance while remaining fully compliant with legal and regulatory requirements\cite{ref1,ref2}.

Currently, in the context of recommendation systems for cross-border data sharing and collaborative business operations, there is a growing demand among institutions to train models by sharing user data. However, the various data requestors and providers involved are not entirely trustworthy, as cross-border data sharing carries significant risks and potential vulnerabilities for information leakag\cite{ref3}.
In general, the privacy challenges in cross-border data sharing for business recommendations can be summarized as follows:
First, given the disparities in data protection and privacy legal frameworks across different countries and regions, the direct sharing of raw data in cross-border scenarios raises substantial privacy concerns and may violate compliance regulations.
Second, to deliver personalized recommendations, these systems typically require the collection and processing of vast amounts of user data, including browsing history, purchase records, search habits, and social connections. Such data often contain sensitive user information, such as age, gender, geographical location, and health status. If not adequately protected during cross-border sharing, this data may be exposed to privacy breaches, posing risks at multiple levels—between users and platforms, among users themselves, and across different platforms.

In recent years, with the increasing occurrence of user privacy incidents and growing public awareness of privacy protection, users have become more vigilant in safeguarding their data privacy to prevent personal information from being collected by internet applications. Governments worldwide have also recognized the significance of data privacy and have enacted relevant laws and regulations concerning data security and privacy protection, such as the General Data Protection Regulation (GDPR) \cite{ref4,ref5} and the California Consumer Privacy Act (CCPA)\cite{ref6}, among others. The implementation of these regulations has, to some extent, guaranteed users' data privacy rights, preventing commercial entities from collecting user data without oversight as they once did.

In addition to legal and regulatory measures, researchers have sought to enhance user privacy protection by refining existing algorithms and designing more robust recommendation system architectures [7]. These approaches can be broadly categorized into architecture-based solutions and algorithm-based solutions. Architecture-based solutions aim to minimize the risk of data leakage. However, these methods may still inadvertently expose user data to other parties and often impose high computational demands on local devices. Algorithm-based solutions, on the other hand, modify raw data in such a way that even if the data or model outputs are intercepted by third parties, user privacy remains uncompromised. These methods primarily include data perturbation and homomorphic encryption algorithms. Nevertheless, a major drawback of such approaches is their substantial computational overhead, storage requirements, and communication costs, rendering them suitable only for small-scale recommendation systems.

While existing cross-domain recommendation methods have achieved notable success, they typically rely on a strong assumption that complete or partial user-item interaction data can be accessed across different domains. However, due to commercial competition and privacy protection concerns, this assumption may not hold in real-world scenarios. Privacy and security issues significantly restrict the cross-domain storage and sharing of highly sensitive data (e.g., interaction records), thereby limiting the practical application of cross-border recommendation models. Although some studies have proposed privacy-aware cross-domain recommendation models, most existing federated learning (FL)--based GNN training approaches still adopt federated averaging strategies, which often underperform on highly heterogeneous graph data. Therefore, there remains a critical need to develop a cross-domain recommendation framework that can simultaneously preserve privacy and enhance recommendation quality to meet real-world application requirements.

\begin{figure*}[!thp]
	\centering		\includegraphics[width=0.7\linewidth]{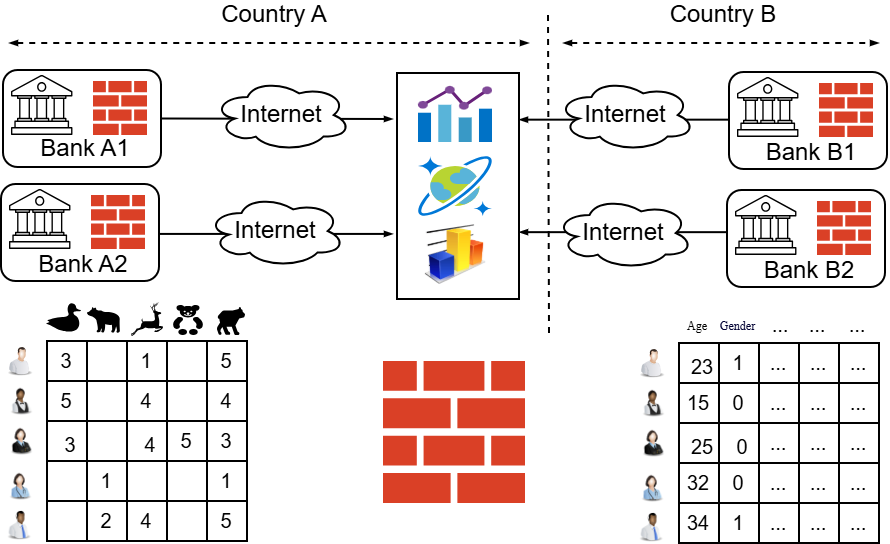}
	  \caption{Cross-border Data Sharing Scenarios for Multi-source Homogeneous Services.}\label{fig1}
\end{figure*}

As is well-known, recommendation systems not only assist users in discovering products that meet their needs but also enable products to efficiently identify potential interested users. From the perspective of cross-border data-sharing business (taking inter-bank cross-border service recommendations as an example), recommendation systems exhibit numerous typical business patterns among geographically distributed banks, as illustrated in Figure \ref{fig1}. For instance: Personalized financial product/service recommendations based on customers' transaction records, behavioral data, and cross-border business demands to enhance customer experience; Integrated service solutions for cross-border loans and investments to streamline operational processes and improve service efficiency; Collaborative financial product promotion among banks to achieve mutual benefits. However, these diverse cross-border recommendation scenarios share a common characteristic: while participating institutions demonstrate high similarity in business operations and feature overlap, they exhibit minimal user overlap. Specifically, between Bank A and Bank B, there exists virtually no intersection in user IDs, yet their user feature spaces show significant commonality.

Addressing the privacy challenges in cross-border data sharing and recommendation scenarios involving multi-source homogeneous data—where participating parties share similar business characteristics with high feature overlap but minimal user overlap—our research motivation is grounded in the data perspective of cross-border operations, focusing on the following key issues: First, given the vast volume of cross-border shared business data, how can we infer implicit relationships between different business datasets and fully exploit their latent features to achieve personalized recommendations, thereby effectively unlocking the potential value of multi-source business data? Second, data fragmentation in recommendation systems may lead to inconsistencies and hinder data accessibility and utilization. Simultaneously, privacy concerns could further restrict data sharing across jurisdictions, exacerbating fragmentation. Thus, how can privacy-preserving computation techniques be leveraged to facilitate secure and efficient cross-border data sharing and circulation without compromising data confidentiality?

Therefore, motivated by the aforementioned research objectives, we propose a Privacy-preserving Federated Recommendation method with Graph Learning (FedGRec), which aims to effectively mitigate privacy leakage risks in such scenarios while facilitating secure data sharing and collaborative applications among multiple parties. In this approach, distributed multi-domain data is leveraged to capture user preferences, and collaborative information from local subgraphs related to users or items is utilized to enhance their representation learning, thereby improving recommendation performance across all subgraphs. Furthermore, by employing graph neural networks, FedGRec exploits high-order connectivity in decentralized user-item interaction graphs to achieve more accurate recommendations. It also automatically identifies and prioritizes target-relevant behaviors while effectively suppressing noise interference from unreliable neighbors, thereby ensuring robust privacy protection for users.

Our contributions are summarized as follows:

\begin{itemize}
\item{We propose FedGRec, a privacy-preserving graph learning method for cross-border federated recommendation, which addresses fragmented data across different jurisdictions in cross-border data element scenarios. This approach not only achieves effective association and complementarity among heterogeneous data sources but also rigorously protects data privacy during cross-border collaborative training.}

\item{We construct dynamic sequential graphs for both target users and candidate items to effectively capture temporally evolving node characteristics. Building upon this, we employ GNNs to decentrally mine high-order connectivity information in user-item graphs, thereby significantly improving recommendation accuracy. Furthermore, through automated filtering of target-relevant behaviors, our method effectively reduces noise interference from unreliable neighbors, ensuring both robustness and reliability of the recommendation system.}

\item{Through extensive experimental evaluations on three representative benchmark datasets, the proposed method not only significantly optimizes model training efficiency but also demonstrates superior performance in terms of model stability.}
\end{itemize}

The remainder of this paper is organized as follows: In Section 2, we provide an in-depth discussion of the research background concerning cross-domain recommendation and its associated privacy challenges. Section 3 focuses on the key definitions and core concepts involved in our proposed methodology, establishing the theoretical foundation for subsequent developments. Section 4 elaborates on the overall architecture of our approach, including its spatiotemporal modules, as well as the federated training methodology for the model. Subsequently, Section 5 validates the feasibility and effectiveness of the proposed method through comprehensive simulation experiments. Finally, Section 6 concludes the paper with a summary of our contributions.

\section{Related works}\label{}

In this section, we provide a concise overview of the current state of cross-border recommendation technologies, while offering a detailed discussion on privacy-preserving research and associated challenges faced by recommendation systems when processing user data in cross-border scenarios.

\subsection{Cross-Domain Recommendation}
Cross-domain recommendation systems aim to address the challenges of information sharing and knowledge transfer across different domains or platforms. By leveraging auxiliary information from external domains, these systems can mitigate data sparsity issues to provide more accurate and personalized recommendation services, thereby improving recommendation quality.
Cross-domain recommendation systems can be categorized into various types based on different classification criteria. According to the number of target domains, they are generally divided into three categories: single-target recommendation, dual-target recommendation, and multi-target recommendation, which are described as follows:

\textbf{Single-target recommendation} primarily focuses on transferring information from one source domain to one target domain. In this scenario, there is typically a data-rich source domain and a data-sparse target domain. The objective is to utilize the abundant information from the source domain to enhance recommendation performance in the target domain\cite{ref7,ref8}. For example:
Transferring user-item rating patterns from the source domain to the target domain to enrich domain knowledge\cite{ref9};
Addressing the non-overlapping recommendation problem based on reviews through attribute alignmen\cite{ref10};
Improving recommendation quality in multiple sparse domains by mining domain-invariant preferences\cite{ref11};
Enabling knowledge transfer for non-overlapping users by exploring joint preferences\cite{ref12}, among others.

\textbf{Dual-Target Recommendation} involves knowledge transfer between two domains, where both domains can potentially benefit from each other. Such recommendation systems focus not only on transferring information from the source domain to the target domain but also consider reverse knowledge transfer from the target domain back to the source domain, or even establishing bidirectional information flow between the two domains. The core challenge of dual-target recommendation lies in learning an effective mapping function that captures the latent relationships between the two domains, thereby facilitating knowledge transfer and recommendation\cite{ref13,ref14}. For instance, cross-domain nonlinear mapping functions—such as those based on linear transformations or Multi-Layer Perceptrons (MLPs)—can be employed to enhance entity embeddings in the target domain\cite{ref15,ref16}.

\textbf{Multi-target recommendation} is inherently more complex, as it involves the transfer of information from multiple source domains to one or more target domains. In such scenarios, there may exist multiple data-rich source domains alongside one or more data-sparse target domains. The objective of multi-target recommendation is to leverage information from all source domains to collectively enhance recommendation performance in the target domain(s). To achieve this goal, it is necessary to design more sophisticated models and algorithms capable of capturing latent relationships across multiple domains and facilitating effective information transfer and integration \cite{ref17,ref18}. For instance, models based on heterogeneous graph embedding or multi-domain collaborative training can effectively handle complex inter-domain relationships and generate recommendations accordingly\cite{ref19,ref20}.

In summary, the core challenge in cross-domain recommendation systems lies in designing an effective transfer method to migrate relevant knowledge from the source (data-rich) domain to the target (data-sparse) domain, thereby improving recommendation accuracy. With recent advances in deep learning, various transfer techniques have emerged in the cross-domain recommendation, including domain adaptatio\cite{ref21}, cross-domain mapping functions \cite{ref22}, deep dual knowledge transfer\cite{ref23}, and graph neural network (GNN)-based methods\cite{ref24,ref25}. However, most existing approaches operate under the assumption that data across all domains are publicly shared, largely neglecting critical privacy concerns associated with user data.

\subsection{Privacy Challenges in Recommendation Systems}

To address the challenges of user privacy protection in recommender systems, researchers have focused on optimizing existing algorithms and designing more scientifically grounded system architectures to ensure robust privacy safeguards. These approaches can be categorized into three main classes: (1) architecture-based solutions, (2) algorithm-driven innovations, and (3) federated learning-based techniques, which are elaborated as follows:

Architecture-Based Data Privacy Protection Schemes aim to minimize data leakage risks through distributed architectural design. Typical approaches include: Distributed Data Storage Mechanisms: Data fragmentation is employed to reduce the potential impact of single-point exposure; Distributed Recommendation Processes: These increase the technical barriers to unauthorized data access. For instance, a user-controlled candidate architecture for data management allows users to autonomously determine the content disclosed to service providers. This framework implements fine-grained access control via API interfaces, permitting only authenticated applications to access configuration data for recommendation computations\cite{ref26}. Another example is a P2P-based distributed recommendation mechanism, which achieves recommendation functionality through localized similarity computation, eliminating the risk of centralized personal data storage on a central server\cite{ref27}. However, these methods still exhibit two significant limitations: Privacy leakage risks persist during data interactions between users; High computational demands are placed on end-user devices.

Algorithm-based privacy-preserving schemes modify raw data through specific transformations to ensure that even if third parties obtain the processed data or model outputs, they cannot infer users' private information. Current mainstream approaches can be categorized into two technical routes: data perturbation and homomorphic encryption.
In the field of data perturbation, researchers achieve privacy protection by designing efficient perturbation mechanisms. A typical method involves adding zero-mean Gaussian noise to user rating data, preventing servers from reconstructing the original information. This technique was first proposed by Agrawal et al., who pioneered the application of additive perturbation in data mining\cite{ref28}. In 2009, McSherry’s team at Microsoft Research introduced differential privacy theory into recommender systems, establishing a crucial theoretical foundation for privacy-preserving research\cite{ref29}. Differential privacy employs perturbation mechanisms on system inputs or outputs, significantly reducing privacy leakage risks. In 2015, Berlioz et al. systematically evaluated the application of differential privacy in matrix factorization, providing an in-depth analysis of the trade-off between privacy protection strength and recommendation accuracy\cite{ref30}.
In contrast, encryption-based solutions offer theoretical advantages in privacy preservation. Homomorphic encryption, as a representative method, enables direct computation on ciphertexts while ensuring that decrypted results are identical to those obtained from plaintext operations. As early as 2002, Canny proposed a homomorphic encryption-based matrix factorization framework, which requires users to encrypt local data using public keys and manage private keys through a distributed key-sharing mechanism\cite{ref31}. Notably, decryption operations require authorization from a majority of online users. However, such methods suffer from high computational complexity, significant storage overhead, and elevated communication costs, limiting their current applicability primarily to small-scale recommendation scenarios.

Federated Learning (FL), first proposed by Google in 2016, is a privacy-preserving machine learning framework whose core idea is to enable distributed collaborative model training without centralized data collection. Unlike traditional machine learning paradigms that rely on centralized data storage and processing, FL coordinates massive end-user devices to participate in model training while keeping raw data locally stored. Instead of sharing raw data, only intermediate computational results (e.g., gradient updates) are exchanged, thereby optimizing the global model while ensuring data privacy. This distributed paradigm not only mitigates privacy risks associated with data centralization but also enhances model performance by leveraging global data. In recent years, FL has achieved significant progress in privacy-preserving recommender systems. Existing research primarily explores the following directions: (1) Fundamental algorithms: Some studies integrate homomorphic encryption with classical federated matrix factorization to mitigate potential privacy leakage risks\cite{ref32}. (2) Model architecture innovations: Researchers have investigated federated graph neural networks (GNNs)\cite{ref33}, and designed federated frameworks based on graph convolutional networks (GCNs) to learn user preference distributions for more accurate recommendations\cite{ref34}. (3) Privacy-enhancing techniques: These include approaches such as local differential privacy with negative sampling\cite{ref35,ref36}, as well as random projection and ternary quantization\cite{ref37}. (4) Cross-domain recommendation scenarios: Proposed solutions include dual-objective vertical federated frameworks\cite{ref38}, federated variational autoencoder models\cite{ref39}, and dual-module architectures\cite{ref40,ref41}. However, most existing studies focus on single privacy-preserving scenarios (either intra-domain or inter-domain), lacking comprehensive solutions for cross-border heterogeneous environments. This study aims to develop a federated recommendation framework that simultaneously addresses both intra-domain and inter-domain privacy protection challenges.

With the growing public awareness of privacy protection, the issue of privacy exposure in cross-border recommendation systems has become increasingly prominent. Meanwhile, alongside the rapid advancement of information technologies, many bottleneck problems that constrain the performance enhancement of privacy-preserving recommendation algorithms are expected to be significantly alleviated. Consequently, privacy-preserving recommendation algorithms are poised to embrace broader development prospects.

\section{Preliminaries}\label{}

\subsection{Problem definition}\label{}

\begin{figure}[!tbp]
\centering
\includegraphics[width=1\columnwidth]{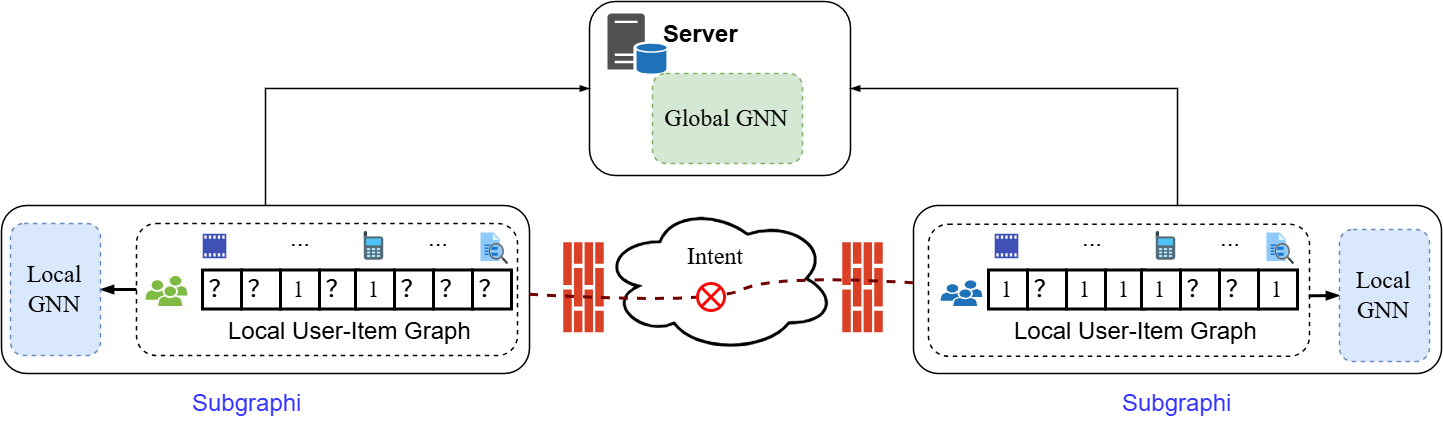}
    \caption{Graph Neural Network-driven learning mechanisms for federated recommendation systems.}
    \label{fig:federalItem}
\end{figure}

In our study, let $U=\left \{u_{1},u_{2},\cdots ,u_{N}\right \}$ denote the set of users and $I=\left \{i_{1},i_{2},\cdots,i_{M}\right \}$ represent the set of items, where $N$ is the number of users and $M$ is the number of items. The $N$ users establish connection relationships with the $M$ items through a cross-border sharing platform. The historical user-item interaction matrix $R=\left \{ r_{ij} \right \}_{N \times M} \in \left \{ 0,1 \right \}^{N \times M}$ is stored locally in the subgraphs of their respective branch institutions rather than being centrally stored on a global server, as illustrated in Figure \ref{fig:federalItem}. Users within the local subgraphs have full access to their historical interactions, while the global server cannot access any raw data residing on the branch clients. Instead, it can only receive processed data transmitted by the clients. Each user's set of interactions constitutes private information that must be protected and must not be disclosed to the server or other users.

From the perspective of graph structure, the user-item interaction graph exhibits a decentralized distribution across various bank branches. Specifically, each branch locally maintains its private subgraph, which comprehensively records all interaction information between users and items within that specific branch. Our primary objective is to accurately process items that have not yet been interacted with by users in any given branch, ultimately generating a personalized list of potentially interesting items for each user $u$.

Given the above description, let $\varepsilon \in \left \{ \left ( u,i,t \right )  \right \}$ denote the observed list of user-item interaction tuples within a time window, where a user $u \in U$ interacts with a set of relevant items $i \in I$ at a given timestamp $t \in T$.

Thus, our problem is formally defined as follows:
During cross-border data sharing, for any given interaction tuple list $\varepsilon$, we aim to maximize user privacy protection while accurately predicting the likelihood of user-item interactions.

\begin{equation}
   \widehat{r}_{ui} =f\left ( u,i,E;\Theta  \right ) =F\left ( u,i,G_{u,t}^{\left ( k \right ) }, G_{i,t}^{\left ( k-1 \right ) };\Phi \right ) 
    \label{eq:interaction}
\end{equation}

where $\Theta$ denotes the network parameters.

\subsection{Multi-source Homogeneous Data}\label{}

\begin{figure}[!tbp]
\centering
\includegraphics[width=1\columnwidth]{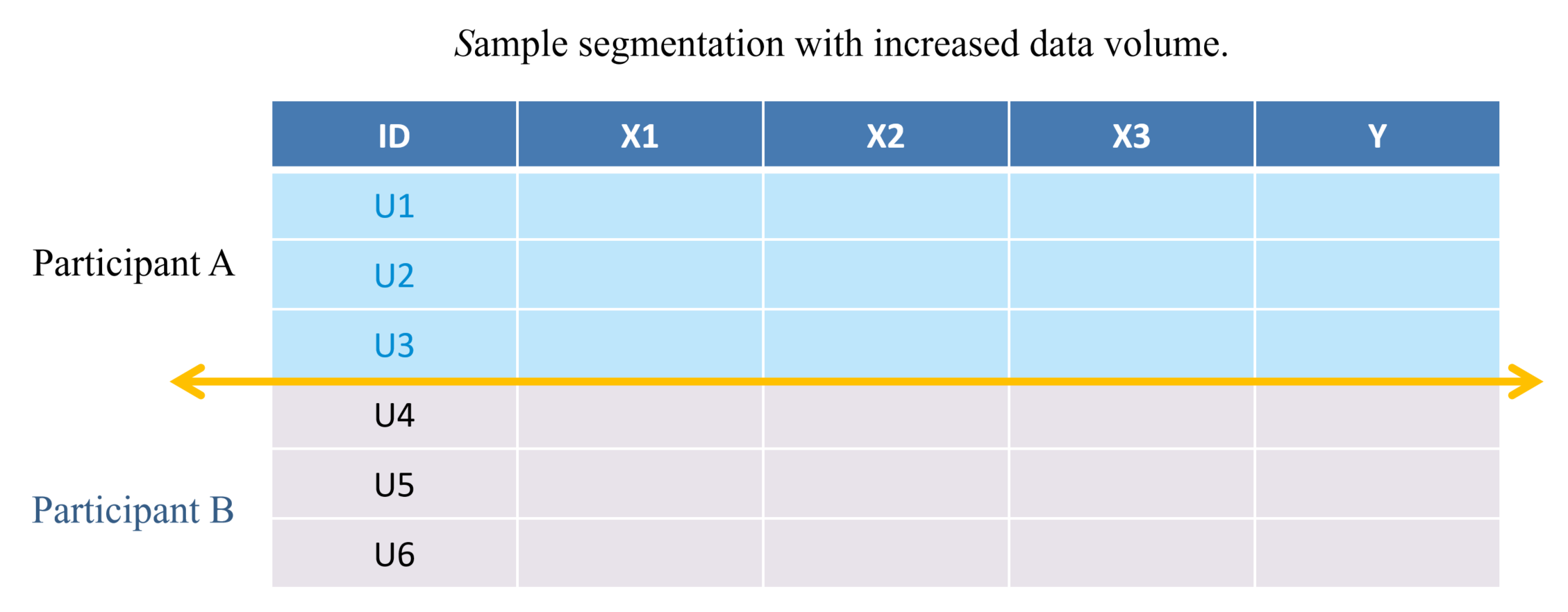}
    \caption{Characteristics of Cross-Border Homogeneous Business Scenarios.}
    \label{fig:Characteristics}
\end{figure}

Multi-source homogeneous data refers to data in cross-border sharing scenarios where participating entities (e.g., enterprises, institutions, or data holders) exhibit high similarity in business models, data types, or feature spaces, but have low overlap in user groups or data samples, as illustrated in Figure \ref{fig:Characteristics}.

Specifically, this scenario satisfies the following two core characteristics:

\begin{itemize}
\item{\textit{Business/Data Homogeneity:} The participating entities share similar business logic (e.g., all are financial institutions). Their data feature spaces exhibit substantial overlap (e.g., all contain homogeneous features such as user profiles and transaction records).}

\item{\textit{Low User/Sample Overlap:} The user groups served or data samples held by different entities have limited intersection (e.g., bank customers from different regions or countries). Individual user data typically resides within a single entity, resulting in weak direct correlations across entities.}
\end{itemize}

Consider a federated learning system with $K$ participants (clients). Each participant $k$ possesses a local dataset $D_{k}$ , and the data distribution satisfies the following conditions:

\begin{itemize}
\item{\textit{Homogeneous Feature Space:} All participants share an identical feature space $X$, i.e., $X_{1}= X_{2}=\dots =X_{K}$, while their sample spaces differ.}

\item{\textit{Sample Independence:} The sample sets across participants are mutually disjoint, i.e., $D_{i}\cap D_{j}=\phi $ for any $i\ne j$.}

\item{\textit{Task Consistency:} All participants maintain the same label space Y and prediction task objective, i.e., $Y_{1}= Y_{2}=\dots =Y_{K}$.}
\end{itemize}

This can be mathematically expressed as:

\begin{equation}%\label{eq:crossAttention}
    \begin{aligned}
       D_{k}=\left \{ \left ( x_{i},y_{i} \right )  \right \}_{i}^{\left ( k \right ) } , x_{i} \in X , Y_{i} \in Y ,\\
       \bigcup_{k=1}^{K} D_{k} = D,  D_{i}\cap D_{j}=\phi, (i\ne j)
    \end{aligned}
\end{equation}

Where $n_{k}$ denotes the number of samples held by participant $k$, and $D$ represents the global dataset.

\subsection{Implicit Relationships}\label{}
It is commonly assumed that connected users may share similar preferences driven by the homophily effect \cite{ref42}. Consequently, most recommendation systems leverage explicit social relationships to enhance recommendation performance. However, beyond explicit relationships, there exist various implicit relationships between users or items that can provide sufficient cues to reveal distinct preferences analogous to explicit relationships, thereby enriching the representation learning of users and recommended items.

\subsubsection{Implicit User Relationships}

Empirical observations indicate that correlated users exhibit preference similarity. However, in most cases, explicit social relationships are sparse and biased, significantly limiting the effective utilization of social information. To address this, we extend from first-order to higher-order relationships to uncover latent implicit associations among users. Specifically, we hypothesize that users who share a substantial number of common interaction followers are likely to exhibit implicit connections based on preference similarity. Formally, for a given user $i$, we define their implicit relationships $H_{U}(i)$ as follows:

\begin{equation}
H_{U}(i)=\left \{k \mid \left \| s_{ji} =1 \cap  s_{jk} =1 , j \in  U \right \| \ge \tau \right \} 
\end{equation}

Here, $s_{ji} =1$ denotes that user $j$ explicitly follows user $i$, $\left \| \cdot  \right \| $ represents the cardinality of the set, $\tau$ indicates the cutoff threshold, and a larger value of $\tau$ implies that the implicit relationship requires more shared followers.

\subsubsection{Implicit Item Relationships}

Traditional item-based collaborative filtering methods employ item similarity metrics to predict user ratings. However, conventional similarity measures such as the Pearson Correlation Coefficient (PCC) typically disregard the scale of rating users. In reality, for cross-border data-sharing product recommendations, the more users who provide similar ratings for two products, the more likely these products are to be genuinely similar - a critical aspect not adequately captured by traditional metrics. To better uncover implicit relationships between products, we propose a refined similarity measurement that incorporates both the number of rating users and their respective rating scales, thereby establishing a more robust implicit product relationship network.

\subsection{Meta-Paths for Personalized Recommendation}\label{}

\begin{figure}[!tbp]
\centering
\includegraphics[width=1\columnwidth]{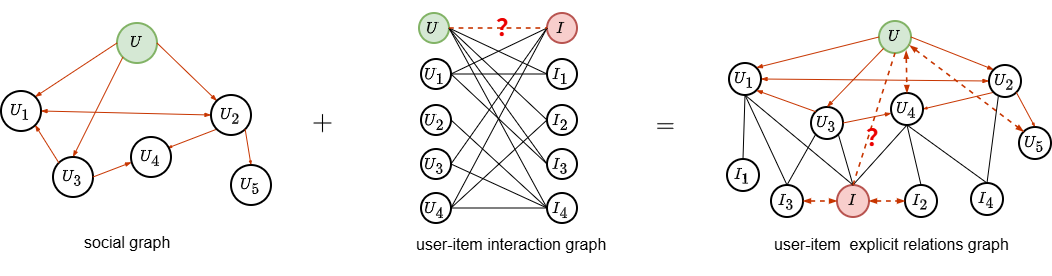}
    \caption{Implicit Relations in Cross-Border Recommendation. The heterogeneous information network is composed of a user-item interaction graph and a social graph. The red dotted lines indicate implicit relationships.}
    \label{fig:Implicit}
\end{figure}

Business context, we map all meaningful user-product interactions into a Heterogeneous Information Network (HIN) \cite{ref43}, as illustrated in Figure \ref{fig:Implicit}. In this network, users and products are represented as nodes, while their interactions—including explicit social connections and implicit relationships—are modeled as different types of edges.

To comprehensively capture user and product preferences, we leverage meta-paths to model the underlying semantic relationships between heterogeneous entities. Specifically, for each user node, we define user-item, user-user, and user-implicit user relationships based on their one-hop neighbors, which directly reflect inherent user behaviors. Additionally, we incorporate two-hop neighbors (e.g., user-user-item and user-implicit user-item) to capture rating behavior similarity along higher-order meta-paths. Similarly, for each item node, we employ corresponding primary meta-paths (e.g., item-user, item-implicit item, and item-implicit item-user) to enable comprehensive representation learning.

\section{Methodology}\label{}

Motivated by the research objectives of this study, which aim to achieve precise user profiling in cross-border data-sharing recommendation scenarios involving multi-source homogeneous data while ensuring compliance with privacy protection requirements for cross-border operations, we propose a model based on FedGRec, with its framework illustrated in Figure 5. Given that most data in recommendation systems exhibit distinct graph-structured characteristics, and considering that GNNs excel at capturing node connections and representation learning in graph-structured data, our method leverages message propagation between nodes in GNNs to model dependencies within the graph. Additionally, by utilizing the powerful and systematic architecture of GNNs, we explore multi-hop neighbor relationships, thereby naturally encoding critical collaborative signals through spatiotemporal modules to enhance the representation learning of users and items.

% \begin{figure*}[thp]
%     \centering    
%     \includegraphics[width=0.85\textwidth]{Figures/Fig5.png}
%     \caption{}
%     \label{fig:architecture}
% \end{figure*}

\begin{figure}[!tbp]
\centering
\includegraphics[width=0.85\columnwidth]{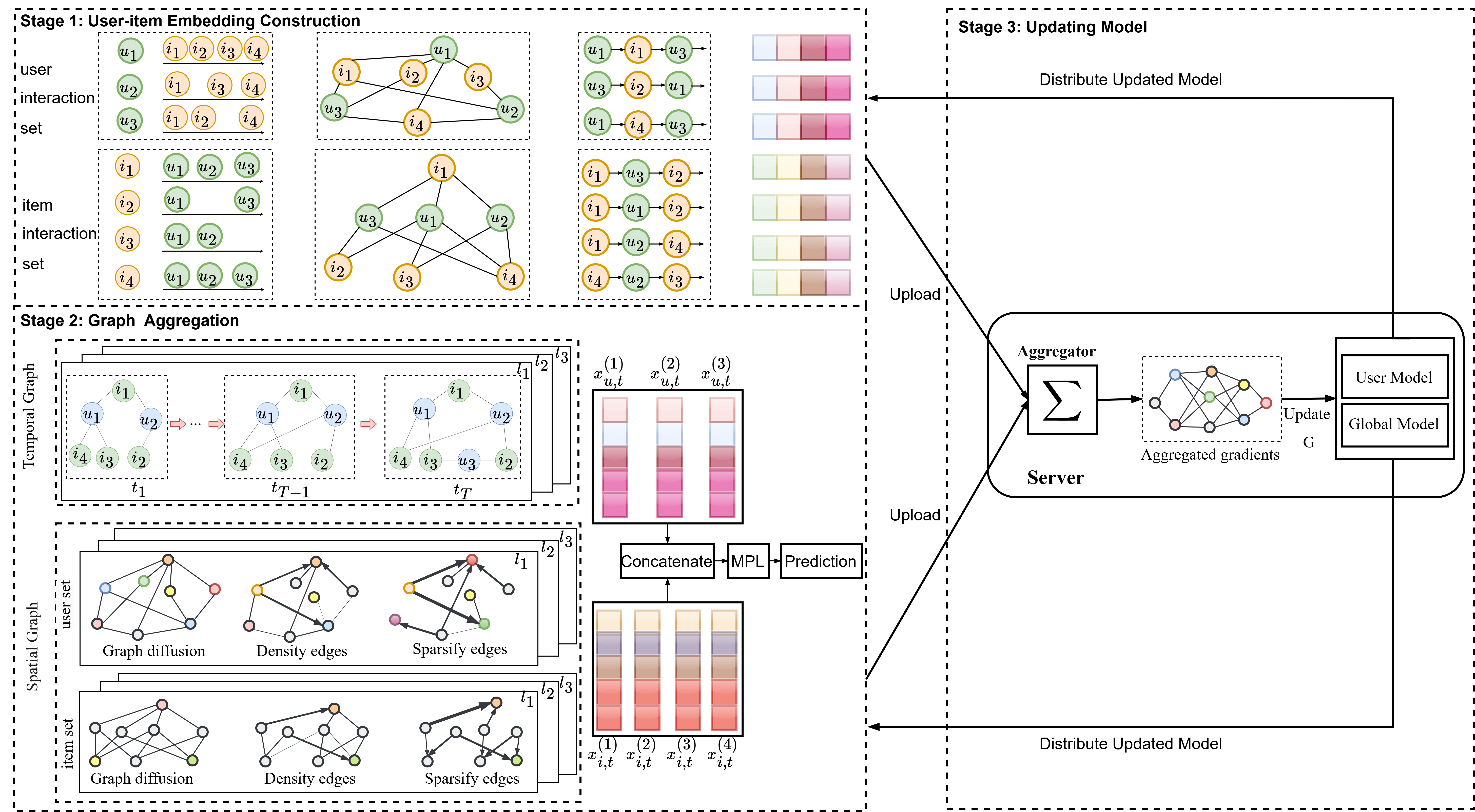}
    \caption{FedGRec architecture: Each subgraph represents a client in federated learning, where Graph Neural Networks (GNNs) are employed to analyze user behaviors, revealing distinct behavioral patterns among local users. The server learns to aggregate data from each client through collaborative learning, thereby integrating user features across subgraphs to achieve coordinated, precise, and dynamic recommendation tasks in heterogeneous environments.}
    \label{fig:architecture}
\end{figure}

In FedGRec, a central server coordinates the model training process, while multiple clients (local private subgraphs from branch institutions) store their sensitive user-item interaction data locally. The recommendation model is constructed by transmitting model parameters between the central server and clients. In cross-border recommendation scenarios, the data requiring protection consists of user-item interactions. Neither the central server nor other users should have access to individual user-item interactions. To address this, we design a spatiotemporal module and three federated learning components: (1) the temporal module captures dynamic variations within user-item interaction sequences; (2) the spatial module characterizes the structural relationships between user (or item) sequences; and (3) federated learning employs secure aggregation \cite{ref43} to protect user privacy across different branches, ensuring that private data from each institution remains localized.

The training process of FedGRec involves three iterative stages of interaction between clients and the central server: user-item embedding, spatiotemporal graph construction, and model updating. In Stage 1, clients generate item-based user embeddings locally. In Stage 2, higher-order embeddings are derived through spatiotemporal graph construction. Once these embeddings are obtained, selected clients compute gradients locally and upload them to the central server for global model updates in Stage 3.

Next, we will provide a detailed introduction to the three underlying modules of the FedGRec model: the temporal module, the spatial module, and the federated learning module.

For each discrete feature field—such as age, gender, category, brand, and $ID$—we represent them as an embedding matrix. By concatenating all feature fields, we obtain the node features of an item, denoted as $f_{item} \in \mathbb{R}^{d_{i}}$. Similarly, $f_{user} \in \mathbb{R}^{d_{u}}$ represents the concatenated embedding vector of the fields in the user category.

\subsection{Temporal Module}\label{}
This module is designed to encode behavioral sequences with temporal information and dependencies. In the temporal module, nodes at each layer are arranged chronologically, reflecting the evolution of user preferences and item popularity over time. To capture the dynamic nature of user-item interaction sequences, we integrate sequence modeling as part of the GNN framework.

At time $t$, for the interaction sequence $\left ( u,i,t \right ) $ between the user and the item, then the historical behavior sequence $B_{j,T}^{u}=\left \{ b_{1}^{u},b_{2}^{u},\dots,b_{T-1}^{u},b_{T}^{u} \right \} $ of the user $u$, where $j=1,2,\dots,T-1,T$,T represents the length of the user behavior sequence;b represents the $i-th$ historical behavior of the user $u$ and may include various auxiliary information, $b_{i}^{u}=\left \{ s_{i,1}^{u},s_{i,2}^{u},\dots,s_{i,k}^{u}\right \} $, $s_{i,k}^{u}$ represents the kth auxiliary information of the $i-th$ behavior of the user $u$, which is generally the $ID$, category, time of occurrence of the behavior, etc.

For a sequence of user's historical behaviours $B_{j,T}^{u}$, we convert each of the user's behaviours into a dense vector: 
\begin{equation}
   e_{j,\tau }^{u} =concat\left ( embedding\left ( f_{item_{j}},f_{item_{\tau}},b_{j,\tau }^{u} \right )  \right ) 
    \label{eq:usersequence}
\end{equation}

By sending each interacting item into the embedding layer along with the time decay in the sequence, the embedded sequence of user behavior can be expressed as: $E_{u} =\left \{ e_{1}^{u},e_{2}^{u},e_{3}^{u},\dots,e_{T}^{u} \right \} $

Similarly, for item $i$'s historical behavior sequence $B_{j,T}^{i} =\left \{ b_{1}^{i},b_{2}^{i},b_{3}^{i},\dots,b_{T}^{i} \right \} $, where $j=,1,2,\dots,T-1,T$, $T$ denotes the length of the item's behavior sequence, each element $b_{j}^{i}$ represents the $j-th$ historical behavior of item $i$ and may incorporate multiple side information features. Formally, we represent each behavior as $b_{j}^{i}=\left \{ s_{j,1}^{i},s_{j,2}^{i},\dots,s_{j,k}^{i}\right \} $, where $s_{j,k}^{i}$ denotes the $k-th$ side information feature associated with the $j-th$ behavior of item $i$.

For the historical behavior sequence $B_{j,T}^{i}$ of an item, we convert each behavior of the item into a dense vector:

\begin{equation}
   e_{j,\tau }^{i} =concat\left ( embedding\left ( f_{item_{j}},f_{item_{\tau}},b_{j,\tau }^{i} \right )  \right )
    \label{eq:usersequence}
\end{equation}

By sending each interacting user into the embedding layer along with the time decay in the sequence, the embedded sequence of the behavior of the item can be expressed as $ E_{i} =\left \{ e_{1}^{i},e_{2}^{i},e_{3}^{i},\dots,e_{T}^{i} \right \} $

% \begin{equation}
  
%     \label{eq:itemsequence}
% \end{equation}

To do this, we use the resulting embeddings as zero-level input to time perception: $X_{u,t}^{\left ( 0 \right )}=e_{j,\tau}^{u}$, $X_{i,t}^{\left ( 0 \right )}=e_{j,\tau}^{i}$.

\begin{figure}[!tbp]
\centering
\includegraphics[width=0.65\columnwidth]{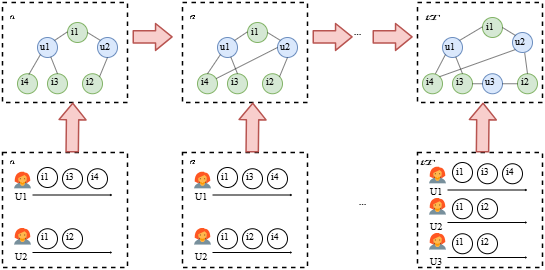}
    \caption{Schematic diagram of time-sliced user-item interaction graph construction. Each dashed box corresponds to a time slice. Note that the number of users/items varies across different time slices, as not all users and items appear in every time slice.}
    \label{fig:Temporal}
\end{figure}

For each time slice, we fully leverage the rich historical interaction information between users and items to construct a global user-item interaction graph, where all users and items appearing in that slice are represented as nodes. This approach enables us to extract knowledge from all user-item interactions, thereby facilitating the learning of user and item representations. The construction process of time-sliced user-item interaction graphs is illustrated in Figure \ref{fig:Temporal}.

Given the constructed sequence of user-item interaction graphs, we employ time-sliced graph neural networks to propagate node representations across different time slices, thereby obtaining temporal user and item representations for each slice.

\subsection{spatial Module}\label{}

In the practical implementation of cross-border data-sharing scenarios, noisy neighbor nodes may exist, whose interests or target audiences are irrelevant to the focal node. To mitigate the interference caused by unreliable nodes, we employ the methodology proposed in \cite{ref44} to activate relevant nodes within behavioral sequences. Specifically, we leverage a multi-head attention mechanism to capture diverse interests or audience preferences \cite{ref45}, formally defined as follows:

\begin{equation}
  h_{i}^{'} = \parallel_{n=1}^{N}\sigma \left ( \sum_{v_{j} \in N\left ( v_{i} \right ) } \alpha_{ij}^{\left ( k \right ) } W^{\left ( k \right ) } h_{j}  \right ) 
    \label{eq:attention}
\end{equation}

Where $\parallel$ represents the splicing operation, $N$ represents the number of repeated node-level attention operations, $\alpha_{ij}^{\left ( k \right ) }$ is the weight coefficient calculated by the $k-th$ attention mechanism, and $W^{\left ( k \right ) }$ is the corresponding learning parameter.

From a spatial perspective, to obtain node representation learning with stronger expressive capabilities, we expect nodes to be able to aggregate messages from not only neighbors (1-order) but also from k-order when updating their feature representation vectors. Messages from all neighbors within. The aggregation method of its nodes is as follows \cite{ref46}:

\begin{equation}%\label{eq:crossAttention}
    \begin{aligned}
     mes_{k}^{\left ( l \right ) } = MES_{u}^{\left ( l,normal \right ) }\left ( \left \{ \left ( h_{u}^{\left ( l-1 \right ) },e_{uv} \right ) \mid u \in Q_{v,G}^{\left ( k,t \right ) } \right \}  \right )    \\+ \sum_{c \in C} \frac{1}{\left | C \right | } \sum_{\left ( i,j  \right )  \in  E\left ( Q_{v,G}^{\left ( k,t \right ) } \right )}e_{ij}\\
     h_{v}^{\left ( l,k\right ) } = UPD_{k}^{\left ( l \right ) }\left (mes_{v}^{\left ( l,k\right ) } , h_{v}^{\left ( l-1\right ) }\right ) \\
      h_{v}^{\left ( l \right ) } = COMBINE^{\left ( l \right )} \left ( \left \{h_{v}^{\left ( l,k \right )} \mid k=1,2,\dots,K \right \}  \right )
    \end{aligned}
\end{equation}

% \begin{equation}\label{eq:motivation1}
%     \begin{split}
%      mes_{k}^{\left ( l \right ) } = MES_{u}^{\left ( l,normal \right ) }\left ( \left \{ \left ( h_{u}^{\left ( l-1 \right ) },e_{uv} \right ) \mid u \in Q_{v,G}^{\left ( k,t \right ) } \right \}  \right )    \\ 
%    + \sum_{c \in C} \frac{1}{\left | C \right | } \sum_{\left ( i,j  \right )  \in  E\left ( Q_{v,G}^{\left ( k,t \right ) } \right )}e_{ij}
%     \end{split}
% \end{equation}

% \begin{equation}
%     h_{v}^{\left ( l,k\right ) } = UPD_{k}^{\left ( l \right ) }\left (mes_{v}^{\left ( l,k\right ) } , h_{v}^{\left ( l-1\right ) }\right ) 
%     \label{eq:attention}
% \end{equation}

% \begin{equation}
%    h_{v}^{\left ( l \right ) } = COMBINE^{\left ( l \right )} \left ( \left \{h_{v}^{\left ( l,k \right )} \mid k=1,2,\dots,K \right \}  \right )
%     \label{eq:attention}
% \end{equation}

Where, $mes_{k}^{\left ( l \right ) }$ represents an aggregation of messages from $Q_{v,G}^{\left ( k,t \right ) }$, and $h_{v}^{\left ( l,k \right ) }$ represents the representation of node $v$ generated based on its previous round characteristics and aggregation of messages from $k-hop$ neighbors. $MES_{u}^{\left ( l,normal \right ) }$ represents the message function of the original GNN model, and $C$ represents the connected component of $Q_{v,G}^{\left ( k,t \right ) }$. 

According to this aggregation method, for the $k+1$ level of node $u$, $x_{u}^{\left ( k \right ) }$, and $\left \{ x_{i}^{\left ( k\right ) } \mid i \in N_{u} \right \}$ are used as input features, then the output feature $x_{u}^{\left ( k+1 \right ) }$ can be abstracted as:

\begin{equation}
   x_{u}^{\left ( k+1 \right ) } = f_{Agg} \left ( x_{u}^{\left ( k \right )}  \mid \left \{x_{i}^{\left ( k \right )} \mid i \in N_{u} \right \}  \right )
    \label{eq:Agg}
\end{equation}

To this end, if a given user's behavior hides the state sequences $\left \{ h_{i,\tau} \mid \left ( i,\tau   \right ) \in B_{j,T}^{u} \right \}$ and $\left \{ h_{u,\tau} \mid \left ( u,\tau   \right ) \in B_{j,T}^{i} \right \} $, after time-sensing sequence encoding, to be able to superimpose the dual attention of time-sensing sequence encoding layers and target preferences, the embeddings obtained at each layer are combined to form The final representation of the user (or item)\cite{ref44}:

\begin{equation}
      \widehat{X}_{u,t} =\frac{1}{K_{u}}\sum_{k=1}^{K_{u}} {x}_{u,t}^{\left ( k \right ) } ;    \widehat{X}_{i,t} =\frac{1}{K_{i}}\sum_{k=1}^{K_{i}} {x}_{i,t}^{\left ( k \right ) } 
    \label{eq:attention}
\end{equation}

Where $K_{u}$ and $K_{i}$ denote the number of layers for user $u$ and item $i$, respectively.

Given the interaction triad $\left ( u,i,t \right )$, the likelihood of user-item interaction can be predicted:

\begin{equation}
    \widehat{r}_{ui} = MLP\left (\left [ e_{u,t}; e_{i,t};\widehat{X}_{u,t};\widehat{X}_{i,t} \right ]  \right ) 
    \label{eq:interaction}
\end{equation}

Where $MLP\left ( . \right )$ denotes the MLP layer.

\subsection{FL Training}\label{}

In the federated learning framework, there are primarily two types of entities: clients and a central server. The clients refer to data holders, which are different branch institutions in this context. The models trained and stored on each client are termed local models. The server aggregates these local models uploaded by each client to obtain a global model. Specifically, assuming there are $K$ clients (i.e., branch institutions), each client possesses a local private subgraph, where $D_{k}$ denotes the data owned by client $k$, and $W_{k}$ represents the local model trained on client $k$. $W_{G}$ signifies the global model maintained on the server.

\begin{figure}[!tbp]
\centering
\includegraphics[width=0.65\columnwidth]{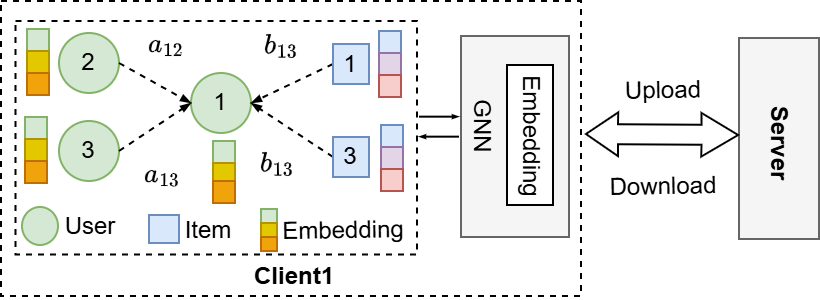}
    \caption{The local GNN derives the embedding representation of node $u_{1}$ through an aggregation mechanism that incorporates the embeddings of its adjacent nodes.}
    \label{fig:Spatio}
\end{figure}

It is noteworthy that all original datasets $D_{k}$, which collectively constitute the complete dataset $D=\left \{ D_{1},D_{2},\dots  D_{K} \right \} $, are exclusively utilized for training their respective local models. These raw datasets are neither uploaded to external servers nor disclosed to any enterprises without explicit user authorization. This approach ensures strict compliance with commercial regulations and user privacy protection policies while simultaneously enhancing the accuracy of recommendation services. Against this backdrop, the training workflow of our model can be summarized through the following key steps:

As illustrated in Figure \ref{fig:Spatio}, each branch institution constructs a user-item graph using local source data and automatically adjusts edge weights in the local graph through an attention mechanism. After connecting user and item latent factors, the results are fed into a Multi-Layer Perceptron (MLP) for rating prediction. Upon completion of local training, each client transmits model parameters and training loss to the central server for aggregation.

During the execution of the prediction task, we define the $\ell_{D_{k}}\left ( W_{k}^{t} \right ) $ function Loss for client $C_{k}$ in the $t-th$ iteration as follows:

\begin{equation}
    \ell_{D_{k}}\left ( W_{k}^{t} \right ) =\frac{1}{2\left |\Theta \right | } \sum_{i,j \in \Theta}\left (r_{ij}^{*}-r_{ij}\right )^{2}  
    \label{eq:interaction}
\end{equation}

Where $W_{k}$ is the model parameter, $\left |\Theta \right | $ is the known number of ratings, $r_{ij}^{*}$ is the predicted rating, and $r_{ij}$ is the known true rating. 

Based on the global model $W_{G}^{t}$ in the $t-th$ iteration, that is, each client uses the local data it holds to train and update the parameters of the local model accordingly. For client $k$, the update formula for the $t-th$ iteration is:

\begin{equation}
    W_{k}^{t} = W_{k}^{t} -\gamma \nabla  \ell_{D_{k}}\left ( W_{k}^{t} \right )
    \label{eq:interaction}
\end{equation}

Where $\gamma$ is a preset learning rate and the updated local model parameters and loss function are uploaded to the central server $S$ for parameter aggregation.

For this purpose, the server aggregates the model parameters uploaded from each client, which in turn generates updated global model parameters $W_{G}^{t+1}$, which are then sent to each client. The commonly used weighted average aggregation formula is as follows \cite{ref47}:

\begin{equation}
    W_{G}^{t+1} = \sum_{k=1}^{K} \frac{\left | D_{k} \right | }{ {\textstyle \sum_{k=1}^{K}}\left | D_{k} \right |  } W_{k}^{t}
    \label{eq:interaction}
\end{equation}

Where $\left | D_{k} \right |$ denotes the number of samples from client $k$.

Algorithm 1 comprehensively delineates the complete workflow of FedGRec, a federated learning -based graph neural network training framework. The computational complexity of FedGRec is O(n²), primarily attributed to the nested local iterations within each global communication round. During each global iteration, clients first retrieve the latest global model parameters from the central server, subsequently perform L steps of local training, and finally upload their locally optimized model parameters back to the server for aggregation.

\section{Experiments}\label{}

Based on the proposed methodology, we design a comprehensive set of experiments closely aligned with the research objectives of this study. The primary focus lies in evaluating the model's usability and performance. To this end, we conduct empirical studies on three public datasets to address the following four key research questions (RQs):

\begin{itemize}
\item  \textbf{RQ1}: How can the privacy-preserving capabilities of FedGRec be validated under varying privacy protection requirements? (Section \ref{RQ1})

\item  \textbf{RQ2}: How does FedGRec perform in comparison to state-of-the-art methods? (Section \ref{RQ2})

\item  \textbf{RQ3}: What is the impact of different components or modules on the performance of the FedGRec model? (Section \ref{RQ3})

\item  \textbf{RQ4}: How do different parameters influence the FedGRec model? (Section \ref{RQ4})
\end{itemize}

\subsection{Experimental Setups}\label{}

\subsubsection{Datasets}\label{}
To comprehensively evaluate the model performance, we selected representative benchmark datasets from LightGCN \cite{ref48}, namely Gowalla, Yelp2018, and Amazon-Book. To mitigate data sparsity issues, we filtered out users with fewer than 5 interactions during preprocessing. The datasets were then randomly partitioned into training (80\%), validation (10\%), and test sets (10\%), with detailed statistical characteristics summarized in Table 1. Following the evaluation protocol of LightGCN, we further sampled 10\% of interaction records from each training set exclusively for hyperparameter fine-tuning and early stopping implementation. This rigorous methodology not only ensures the reliability of evaluation outcomes but also enhances the model's generalization capability and practical utility.

\begin{table}[!tb] 
\centering
\caption{Statistical analysis of the experimental data.}
\scalebox{1.0}{
    \begin{tabular}{lccr}
      \hline%\toprule[1pt]
        Dataset         &Gowalla &Amazon-Book    &Yelp2018\\ \midrule
		$\#$user        &29,858  &52,643         &31,668\\ 
        $\#$Item        &40,981  &91,599        &38,048\\
        $\#$Interaction &1,027,370  &2,984,108     &1,561,406\\ 
        Density         &$8.4×10^{-4}$  &$6.2×10^{-4}$  &$1.3×10^{-4}$\\ 
      \hline%\bottomrule[1pt]
    \end{tabular}  
    }
    \vspace{-1.5em}
\end{table}

However, during the experiments on the Gowalla dataset, we observed an intriguing phenomenon: the performance on the validation set continued to improve, while the test set performance exhibited a declining trend. We hypothesize that this discrepancy may stem from a significant distributional shift between the validation and test sets. To address this issue, we opted to re-partition the Gowalla dataset by adopting an 80:20 split ratio to redefine the training and test sets, while independently constructing a validation set from the training portion. This adjustment was implemented to ensure more accurate and reliable experimental results.

\subsubsection{Evaluation Metrics}\label{}
To comprehensively evaluate the performance of various recommendation algorithms, we employed three widely recognized predictive evaluation metrics: Root Mean Square Error (RMSE), Mean Absolute Error (MAE), and Mean Absolute Percentage Error (MAPE). Lower values of these metrics indicate higher prediction accuracy. To ensure the stability and reliability of the results, we conducted 10 repeated experiments and reported the average performance on the test dataset corresponding to the best-performing epoch in the validation set.

For the evaluation, the model generates a ranked list of items for each user by sorting all items with which the user has not previously interacted. Following the evaluation protocol of LightGCN, we adopted two widely used metrics: Recall and Normalized Discounted Cumulative Gain (NDCG), with the number of recommended items set to $K = 20$ by default. In other words, we utilized Recall@20 and NDCG@20 to assess the top 20 items in each ranked list.

\subsubsection{Baselines}\label{}

To evaluate the performance of the proposed FedGRec against baseline recommendations, the following methods were selected for comparison:

\begin{itemize}
\item  \textbf{Local Training}: A single local client was chosen to assess the feasibility and effectiveness of the proposed method.

\item  \textbf{LightGCN \cite{ref49}}: A competitive GNN-based recommendation method that eliminates unnecessary complexity in collaborative filtering. It consists of two key components—lightweight graph convolution and layer combination—making it more concise and better suited for recommendations.

\item  \textbf{BiTGCF \cite{ref50}}: A transfer learning approach for cross-domain recommendation that integrates high-order feature propagation in graph structures with transfer learning. The method comprises two core modules: a feature propagation module and a feature transfer module.

\item  \textbf{FedCDR \cite{ref51}}: A personalized federated learning framework designed for privacy-preserving cross-domain rating prediction, consisting of a rating prediction model and a cross-domain recommendation model. It employs a simple matrix factorization (MF)-based recommendation model as its backbone.

\item  \textbf{P2FCDR \cite{ref52}}: A peer-to-peer federated architecture that ensures local data storage and privacy protection for business partners. By leveraging the similarity between intra-domain and cross-domain embeddings, a gated selection vector is developed to guide information fusion for more accurate bidirectional transfer.

\item  \textbf{PPCDR \cite{ref53}}: A federated graph learning method for privacy-preserving cross-domain recommendations using distributed multi-domain data. It models both global preferences across multiple domains and domain-specific local preferences for a given user, capturing shared and domain-specific user preferences over interacted items.

\end{itemize}

\subsubsection{Parameter Settings}\label{}

We implement the proposed FedGRec model using the PyTorch framework and optimize the loss function with the Adam optimizer \cite{ref54} to determine the optimal hyperparameter configuration. The learning rate is set to 0.001, while the embedding dimension and batch size are fixed at 256 and 128, respectively. The model is trained for 200 epochs. For the user-item bipartite graph, we initialize DropNode with a rate of 25\%, and the implicit neighbor size in the social graph is set to 20. To construct implicit user-item relationships, we select the top 20 users and items based on similarity. To mitigate overfitting, we employ early stopping and normalization techniques. For a fair comparison, all baseline models are carefully fine-tuned on our dataset to achieve their best performance.

\subsection{Experimental Results}\label{}

\subsubsection{Privacy Validation (RQ1)}\label{RQ1}

We systematically analyzed how different approaches address diverse privacy-preservation requirements. Specifically, for each user, a distinct public interaction ratio was configured based on real-world scenarios, representing the proportion of interactions that users were willing to disclose relative to their total interaction behaviors.
In centralized methods, both graph construction and model training were strictly confined to users' publicly shared interactions, preventing access to undisclosed data, which inherently constrained their performance potential. In contrast, federated learning approaches adhered to privacy-preserving principles during graph construction, utilizing only publicly available interactions. However, during model training, federated learning demonstrated a unique advantage by comprehensively integrating and leveraging users' complete interaction data—including non-public interactions—thereby significantly enhancing model performance and generalization capability while preserving user privacy.

\begin{figure}[ht]
    \centering
    \begin{subfigure}[b]{0.3\textwidth}
        \includegraphics[width=\textwidth]{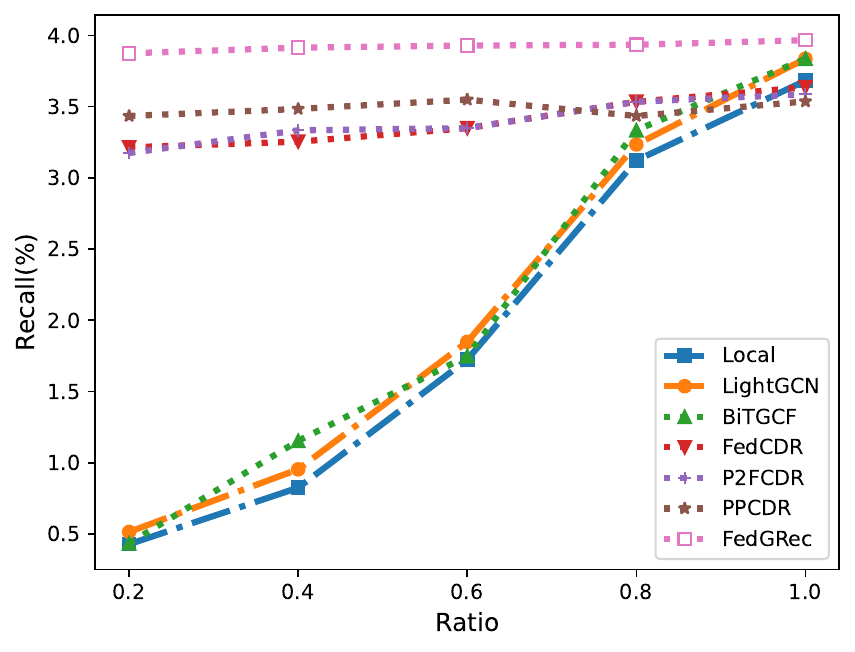}
        \caption{Gowalla}
        \label{fig:sub1}
    \end{subfigure}
    \hfill
    \begin{subfigure}[b]{0.3\textwidth}
        \includegraphics[width=\textwidth]{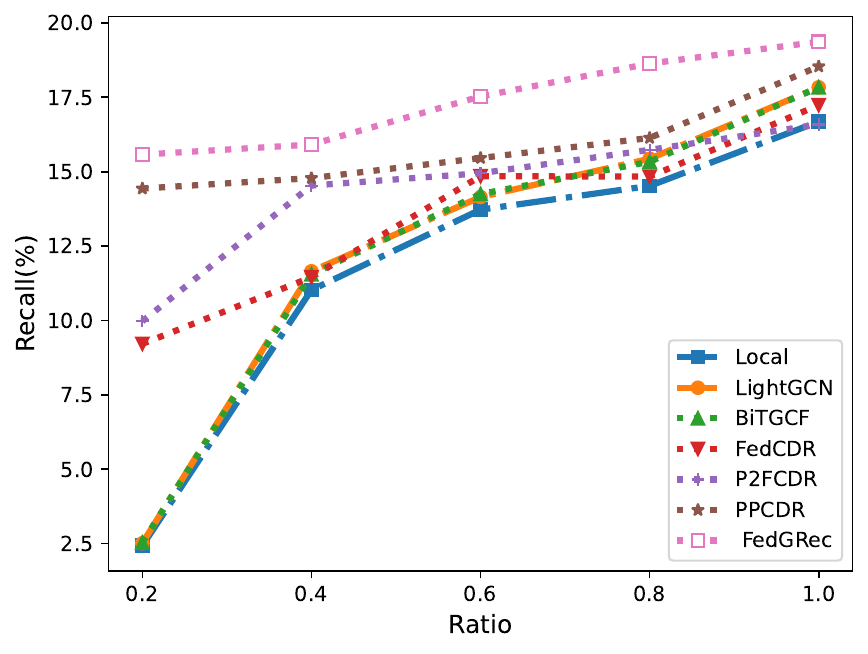}
        \caption{Amazon-Book}
        \label{fig:sub2}
    \end{subfigure}
    \hfill
    \begin{subfigure}[b]{0.3\textwidth}
        \includegraphics[width=\textwidth]{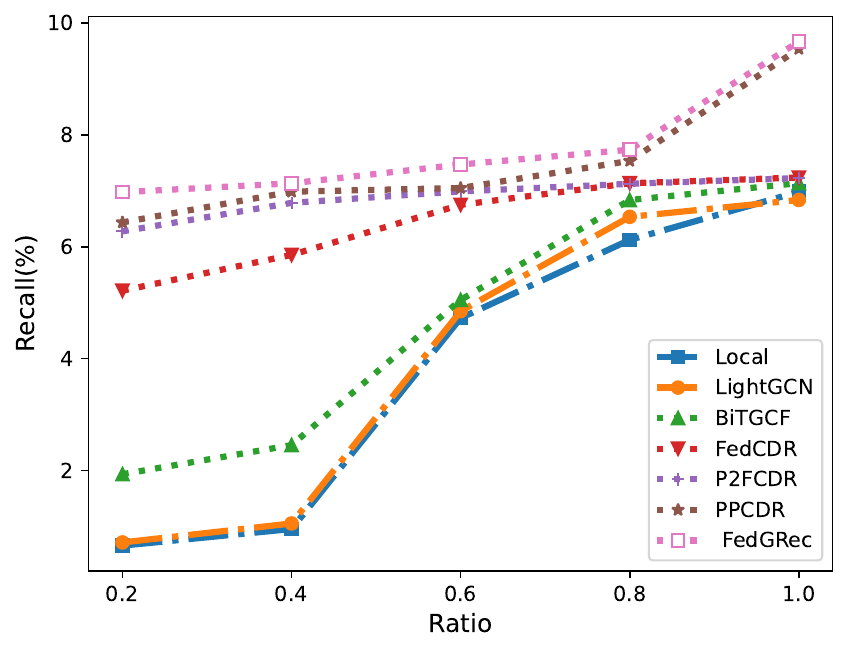}
        \caption{Yelp2018}
        \label{fig:sub3}
    \end{subfigure}
    
    \caption{Performance with different ratios of public interactions (Gowalla, Amazon-Book, Yelp2018)}
    \label{fig:RQ1}
\end{figure}

As illustrated in Figure \ref{fig:RQ1}, comparative experiments conducted on the Gowalla, Amazon-Book, and Yelp2018 datasets yielded the following key observations:
(1) As the public interaction ratio decreased, centralized methods exhibited a pronounced performance decline, whereas FedGRec and other federated learning approaches demonstrated superior robustness, maintaining relatively stable recommendation performance.
(2) Notably, when the public interaction ratio was set below 1, FedGRec consistently outperformed all baselines across all datasets, robustly validating its capability to accommodate varying privacy-preservation demands.

% \begin{figure}[!tbp]
% \centering
% \includegraphics[width=0.95\columnwidth]{Figures/Fig8.png}
%     \caption{Performance with different ratios of public interactions (Gowalla, Amazon-Book, Yelp2018)}
%     \label{fig:RQ1}
% \end{figure}

\subsubsection{Performance Comparison (RQ2)}\label{RQ2}

we first conducted a comprehensive performance comparison between the proposed FedGRec method and several baseline approaches, including both centralized recommendation methods based on aggregated user data storage and privacy-preserving federated learning-based recommendation methods. The detailed comparative results are presented in Table \ref{tab:Performance}. Among these methods, Local, LightGCN, and BiTGCF represent centralized approaches, while FedCDR, P2FCDR, PPCDR, and our FedGRec belong to federated methods. From the table, we can observe the following key findings:

\begin{table}[H] 
\caption{Performance comparison between FedGRec and baseline methods.} 
    \label{tab:Performance}
  \centering
  %\fontsize{8}{10}\selectfont      
	  \resizebox{0.8\columnwidth}{!}{      
        \begin{tabular}{lccccccccccc}
        \toprule
        \rule{0pt}{12pt}
        \multirow{2}{*}{Methods}&
        \multicolumn{2}{c}{Gowalla}&\multicolumn{2}{c}{Yelp2018}&\multicolumn{2}{c}{Amazon-Book}\cr
        \cmidrule(lr){2-3} \cmidrule(lr){4-5} \cmidrule(lr){6-7} 
        &Recall &NDCG  &Recall &NDCG  &Recall &NDCG \cr
        \midrule
		Local	   &0.0328±0.0024	&0.0268±0.0020	&0.0670±0.0028	&0.0376±0.0023	&0.1585±0.0026	&0.1015±0.0012\cr
		LightGCN   &0.0347±0.0041	&0.0313±0.0027	&0.0994±0.0023	&0.0549±0.0014	&0.1878±0.0010	&0.1225±0.0007\cr
		BiTGCF	   &0.0389±0.0034	&0.0349±0.0030	&0.0836±0.0041	&0.0434±0.0024	&0.2017±0.0028	&0.1312±0.0021\cr
		FedCDR	   &0.0324±0.0046  &0.0262±0.0034	&0.0632±0.0039	&0.0350±0.0032	&0.1518±0.0035	&0.0927±0.0025\cr
		P2FCDR	   &0.0338±0.0052	&0.0302±0.0036	&0.0740±0.0044	&0.0389±0.0033	&0.1729±0.0029	&0.1084±0.0013\cr
		PPCDR	   &0.0330±0.0037	&0.0284±0.0028	&0.0773±0.0031	&0.0408±0.0024	&0.1850±0.0026	&0.1189±0.0019\cr
		FedGRec	&{\bf0.0476±0.0041}	&{\bf0.0427±0.25}	&{\bf0.0968±0.0035}	&{\bf0.0513±0.0021}	&{\bf0.1991±0.31}	&{\bf0.1276±0.0018}\cr
        \bottomrule
        \end{tabular}}
\end{table}

(1) Centralized recommendation methods demonstrate superior performance across all datasets, which can be attributed to their capability of leveraging complete user interaction data for recommendation generation. In contrast, federated approaches, while facing challenges of data fragmentation due to privacy preservation requirements, still achieve competitive performance. This convincingly validates the feasibility and effectiveness of generating accurate recommendations without compromising user privacy.

(2) Compared with matrix factorization (MF) based methods or other approaches (e.g., FedCDR), graph neural network (GNN) methods exhibit notable advantages, highlighting GNN's superior capability in recommendation systems. This advantage stems from GNN's ability to significantly enrich and enhance the representation learning of both users and items. Furthermore, we observe that methods incorporating higher-order information from user-item graphs (such as FedGRec and BiTGCF) achieve even better performance. This improvement likely occurs because modeling complex high-order interactions between users and items, as opposed to relying solely on first-order information, can substantially deepen and broaden the learning of user and item representations, thereby improving recommendation accuracy.

(3) Compared with other baseline methods, our proposed FedGRec achieves comparable or even superior performance. Experimental results demonstrate that this method delivers satisfactory recommendation quality while rigorously protecting user privacy. This success benefits from FedGRec's effective integration of high-order information from user-item graphs, which strengthens its representation power. Moreover, our approach features a dual protection mechanism that simultaneously safeguards both rating scores and user-item interaction histories.

Furthermore, we conducted a comprehensive analysis of the overall performance of all models based on the two evaluation metrics described in the experimental setup, namely RMSE (Root Mean Square Error) and MAE (Mean Absolute Error). The actual comparative results across the three datasets are presented in Table \ref{tab:RMSE}.

\begin{table}[!b] 
\caption{Performance of different methods in terms of RMSE and MAE.} 
    \label{tab:RMSE}
  \centering
  %\fontsize{8}{10}\selectfont      
	  \resizebox{0.95\columnwidth}{!}{      
        \begin{tabular}{lccccccccccc}
        \toprule
        \rule{0pt}{12pt}
        \multirow{2}{*}{Methods}&
        \multicolumn{2}{c}{Gowalla}&\multicolumn{2}{c}{Yelp2018}&\multicolumn{2}{c}{Amazon-Book}\cr
        \cmidrule(lr){2-3} \cmidrule(lr){4-5} \cmidrule(lr){6-7} 
        &RMSE &MAE  &RMSE &MAE  &RMSE &MAE \cr
        \midrule
		Local	   &0.8029±0.0023	&1.0541±0.0036		&0.7252±0.0043	&0.9581±0.0069		&0.7829±0.0023	&1.0145±0.0037\cr
        P2FCDR	   &0.7903±0.0039	&1.0707±0.0113		&0.7168±0.0110	&0.9832±0.0135		&0.7885±0.0027	&1.0281±0.0037\cr
        BiTGCF	   &0.8172±0.0041	&1.0826±0.0145		&0.7353±0.0049	&1.0005±0.0072		&0.7947±0.0037	&1.0409±0.0049\cr
        FedCDR	   &0.8045±0.0016	&1.0707±0.0026		&0.7302±0.0054	&0.9957±0.0075		&0.7867±0.0025	&1.0346±0.0032\cr
        LightGCN   &0.8254±0.0032	&1.0789±0.0056		&0.7482±0.0052	&0.9890±0.0076		&0.8037±0.0034	&1.0680±0.0052\cr
        PPCDR	   &0.7813±0.0023	&1.0583±0.0055		&0.7193±0.0075	&0.9717±0.0140		&0.7726±0.0075	&1.0251±0.0048\cr
        FedGRec	   &{\bf0.7654±0.0028}	&{\bf1.0396±0.0046}		&{\bf0.7007±0.0052}	&{\bf0.9498±0.0063}		&{\bf0.7543±0.0023}	&{\bf1.0026±0.0028}\cr
        \bottomrule
        \end{tabular}}
\end{table}

Notably, to ensure the stability and accuracy of the experimental results, each reported outcome was obtained by averaging over 10 independent repetitions of the experiment. In our evaluation, we employed a centralized training approach, where the recommendation model was trained on the complete dataset, serving as a benchmark to assess the performance of the federated learning framework. Generally, centralized training with full data access yields superior model performance compared to federated learning, which relies on partial and distributed data. However, due to stringent privacy protection requirements, centralized training is often impractical in real-world applications.

Against this backdrop, we specifically selected a locally trained model—Client 1—as a case study to validate the effectiveness and feasibility of cross-institutional training. The performance of Client 1 not only demonstrates the viability of model training under limited data conditions but also highlights the potential of optimized cross-border data sharing in enhancing model performance, even under strict privacy constraints.

Meanwhile, as demonstrated in Table \ref{tab:RMSE}, both the models trained using federated learning (FL) approaches and those trained solely on local data underperform compared to the centrally trained model. Notably, the locally trained models exhibit the poorest performance, which can be attributed to data scarcity faced by certain data demand parties and the lack of cross-border data collaboration. Although the FL-based models surpass other baseline methods in evaluation metrics, they still fall short of meeting the standard set by Client 1. This underscores the urgent need to develop an FL-based framework for training recommendation models in cross-border business sharing scenarios.

Among all baseline methods, the graph neural network (GNN)-based recommendation approach significantly outperforms non-GNN methods, strongly validating the unique advantages of GNNs in processing graph-structured data. Compared to the PPCDR method, our proposed FedGRec framework achieves average improvements of 0.0176 in RMSE and 0.0311 in MAE. These results not only confirm FedGRec's robust capability in accurately distinguishing user interaction patterns within cross-border heterogeneous data sharing but also highlight the distinctive benefits of our high-order aggregation strategy in enhancing node representation learning.

\subsubsection{Ablation study (RQ3)}\label{RQ3}

To validate the effectiveness of different modules or components in our proposed FedGRec model on the Gowalla, Amazon-Book, and Yelp2018 datasets, we conducted the following ablation studies:

(1) In the construction of the item hypergraph, we explored two variants: w/o item graph and neighbor. w/o item graph simulates user preferences by averaging the embeddings of interacted items while disregarding the structural information of the item graph itself. w/o neighbor excludes the publicly available interaction information from neighbors when constructing the item hypergraph. (2) To evaluate the variants of user preference modeling, we introduced the w/o attention setting, where the interest attention mechanism was removed and replaced with traditional attention pooling to assess the contribution of the attention mechanism to user preference modeling. (3) To examine the impact of different implicit relations on model performance and verify whether implicit relations provide equally valuable information as explicit social relations, we designed w/o implicit relations. By isolating the influence of implicit relations, we were able to observe their contribution to overall performance.

The comparative results of the ablation studies are presented in Table \ref{tab:variants}. Based on this analysis, we draw the following conclusions:

\begin{table}[!b] 
\caption{Performance on all variants of FedGRec.} 
    \label{tab:variants}
  \centering
  %\fontsize{8}{10}\selectfont      
	  \resizebox{0.9\columnwidth}{!}{      
        \begin{tabular}{lccccccccccc}
        \toprule
        \rule{0pt}{12pt}
        \multirow{2}{*}{Methods}&
        \multicolumn{2}{c}{Gowalla}&\multicolumn{2}{c}{Yelp2018}&\multicolumn{2}{c}{Amazon-Book}\cr
        \cmidrule(lr){2-3} \cmidrule(lr){4-5} \cmidrule(lr){6-7} 
        &Recall &NDCG  &Recall &NDCG  &Recall &NDCG \cr
        \midrule
        w/o item graph	    &0.0328±0.0034	&0.0368±0.0035	&0.0670±0.0028	&0.0376±0.0023	&0.1585±0.0026	&0.0915±0.0012\cr
        w/o neighbor	    &0.0347±0.0031	&0.0313±0.0024	&0.0694±0.0023	&0.0549±0.0014	&0.1878±0.0010	&0.1125±0.0007\cr
        w/o attention	    &0.0389±0.0043	&0.0349±0.0032	&0.0836±0.0041	&0.0434±0.0024	&0.1617±0.0028	&0.1112±0.0021\cr
        w/o implicit user	&0.0424±0.0026	&0.0392±0.0036	&0.0632±0.0039	&0.0350±0.0032	&0.1518±0.0035	&0.1027±0.0025\cr
        w/o implicit item 	&0.0424±0.0026	&0.0392±0.0036	&0.0632±0.0039	&0.0350±0.0032	&0.1518±0.0035	&0.1027±0.0025\cr
        FedGRec	&{\bf0.0476±0.0041}	&{\bf0.0427±0.0025}	&{\bf0.0968 ±0.0035}	&{\bf0.0513±0.0021}	&{\bf0.1991±0.0031}	&{\bf0.1276±0.0018}\cr
        \bottomrule
        \end{tabular}}
\end{table}

(1) Each component in FedGRec plays an indispensable role. Among them, the construction of the item graph and the utilization of neighbors' publicly shared interactions serve as the two core drivers for significant performance improvement. The former accurately distinguishes users' core interests from temporary interests, while the latter leverages behavioral similarities among users to further enhance recommendation effectiveness. (2) Merely constructing an item-tag bipartite graph fails to demonstrate competitive performance advantages, particularly in datasets with the richest user interaction data, where its performance is notably inferior. (3) The introduction of an attention mechanism, which aggregates high-order neighbor information, further improves model performance and significantly strengthens recommendation capabilities. (4) Both implicit user relationships and item relationships positively contribute to recommendation performance, collectively optimizing the overall efficacy of the recommendation system.

\subsubsection{Parameters Sensitivity (RQ4)}\label{RQ4}

\textit{(1) Hyperparameters Sensitivity}

\begin{figure}[ht]
    \centering
    \begin{subfigure}[b]{0.3\textwidth}
        \includegraphics[width=\textwidth]{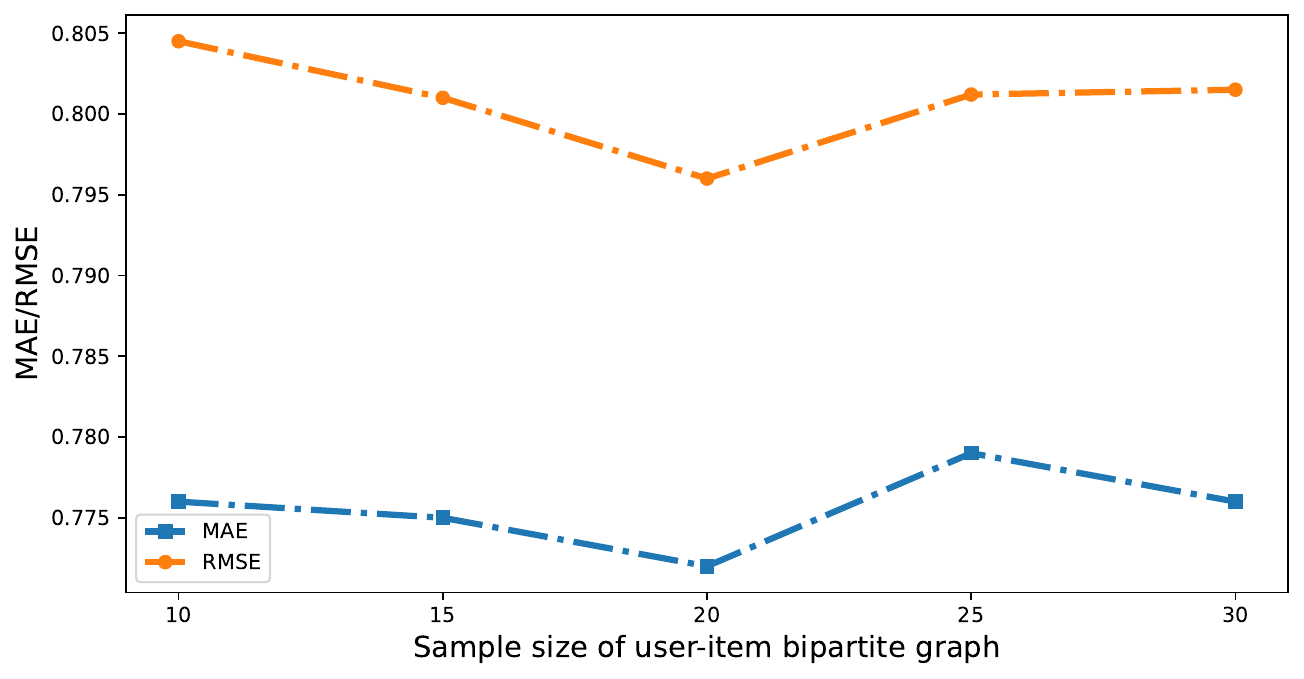}
        \caption{\#rated users/items}
        \label{fig:sub1}
    \end{subfigure}
    \hfill
    \begin{subfigure}[b]{0.3\textwidth}
        \includegraphics[width=\textwidth]{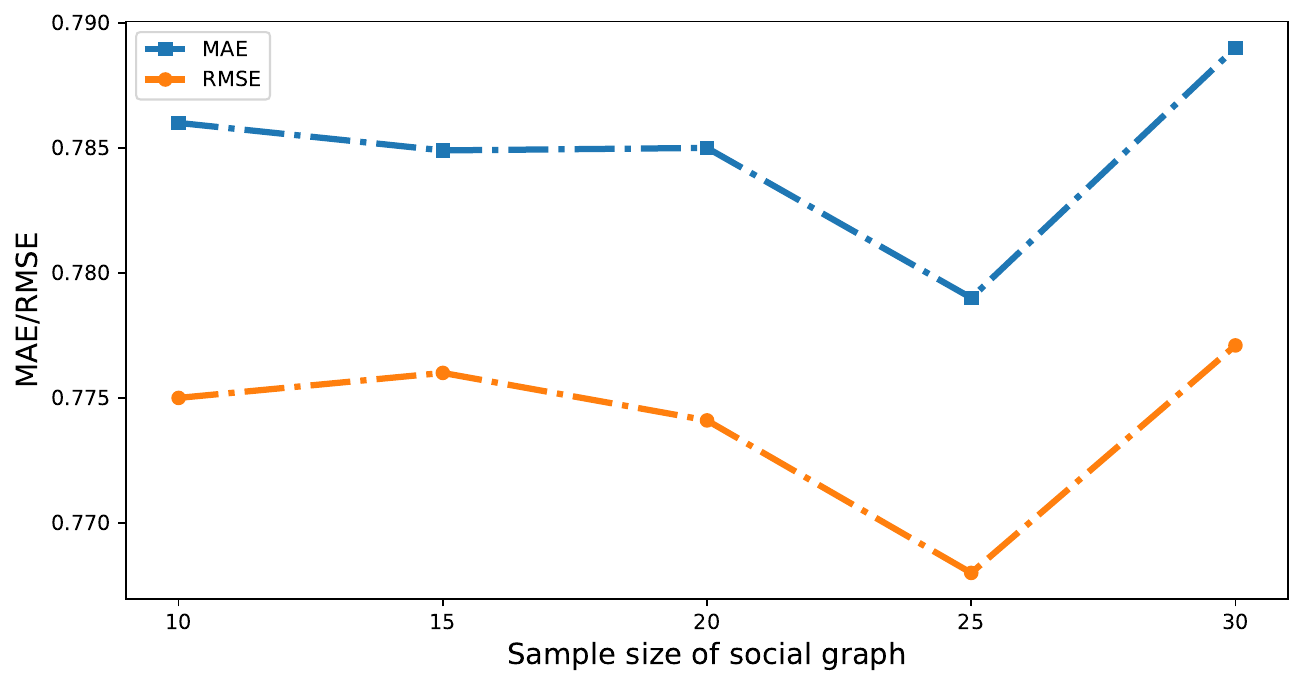}
        \caption{\#friends}
        \label{fig:sub2}
    \end{subfigure}
    \hfill
    \begin{subfigure}[b]{0.3\textwidth}
        \includegraphics[width=\textwidth]{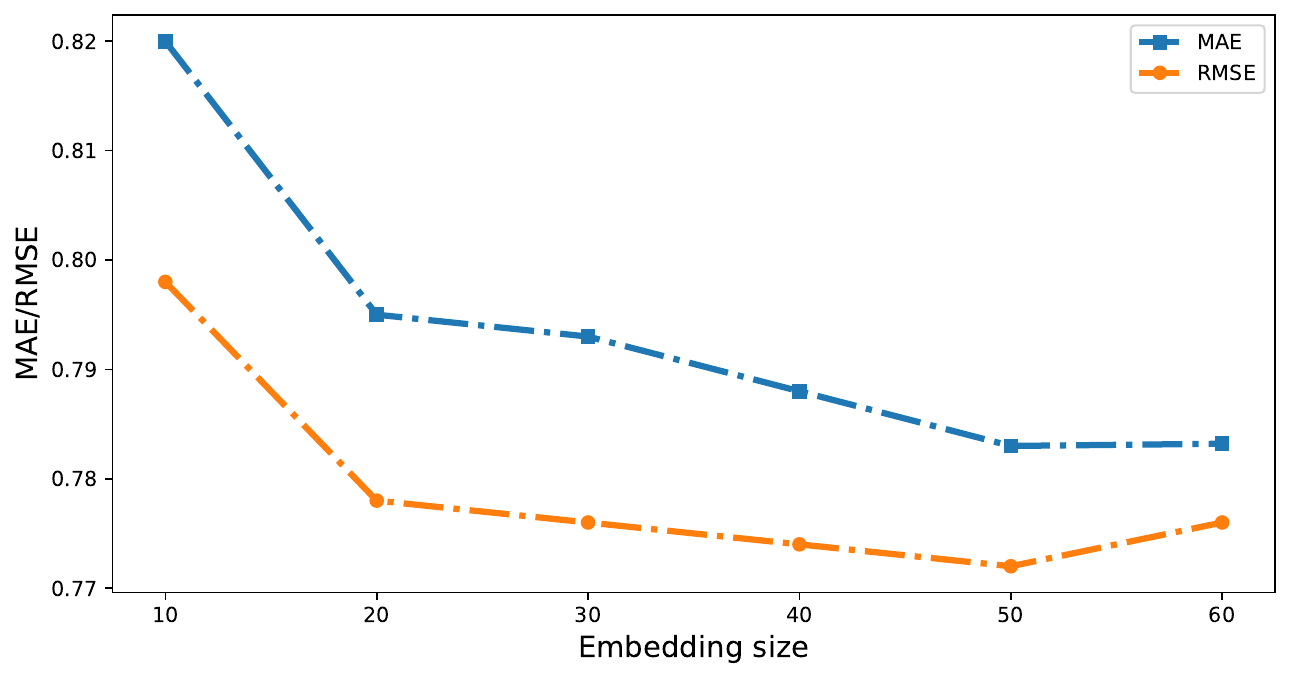}
        \caption{\#embedding size}
        \label{fig:sub3}
    \end{subfigure}
    
    \caption{Performance on Amazon-Book w.r.t different hyper-parameters.}
    \label{fig:hyperparameters}
\end{figure}

In this section, we investigate the performance variations of the proposed model when adjusting several critical hyperparameters, including the embedding dimension, the scale of node deletion in the user-item interaction graph, and the number of neighbors in the social graph. Due to space constraints, our analysis primarily focuses on the performance outcomes observed on the Amazon-Book dataset, as illustrated in Figure \ref{fig:hyperparameters}, while omitting other hyperparameters that exhibit relatively negligible impacts on model performance.

% \begin{figure}[!tbp]
% \centering
% \includegraphics[width=0.95\columnwidth]{Figures/Fig8.png}
%     \caption{Performance on Amazon-Book w.r.t different hyper-parameters.}
%     \label{fig:hyperparameters}
% \end{figure}

Regarding the influence of embedding dimensions, it is noteworthy that the embedding size is directly correlated with the model's capacity. As demonstrated in Figure \ref{fig:sub1}, as the embedding dimension of the deep model increases, the model exhibits overall performance improvements owing to its enhanced expressive power. However, once the embedding dimension surpasses a certain threshold, the marginal performance gains become insignificant, while computational complexity escalates substantially. Consequently, selecting an appropriate embedding dimension is crucial for achieving an optimal balance between model capacity and computational efficiency.

In the neighbor addition/deletion tests, we further investigated the impact of sample size on the performance of node removal operations in both user-item interaction graphs and social graphs, with the corresponding results illustrated in Figures \ref{fig:sub2} and \ref{fig:sub3}. Our observations reveal that model performance improves progressively with increasing sample size, which can be attributed to enhanced sample diversity facilitating more accurate learning of user and item representations. However, when the sample size exceeds a certain threshold, the learned representations begin to exhibit bias and contamination, ultimately exerting detrimental effects on performance while incurring substantial computational overhead. Consequently, selecting an appropriate sample size emerges as a critical factor in balancing prediction accuracy with training efficiency.

\textit{(2) Parameter analyses}

\begin{figure}[!tbp]
    \centering
    \begin{subfigure}[b]{0.3\textwidth}
        \includegraphics[width=\textwidth]{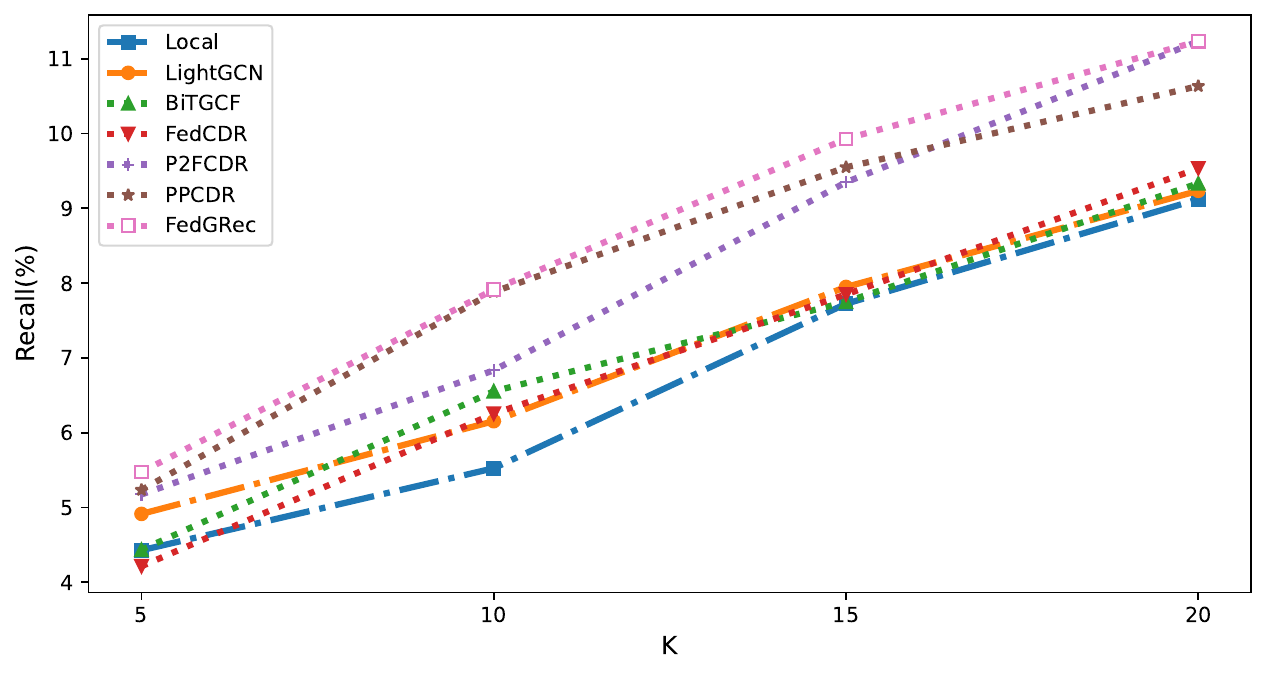}
        \caption{Gowalla}
        \label{fig:sub1}
    \end{subfigure}
    \hfill
    \begin{subfigure}[b]{0.3\textwidth}
        \includegraphics[width=\textwidth]{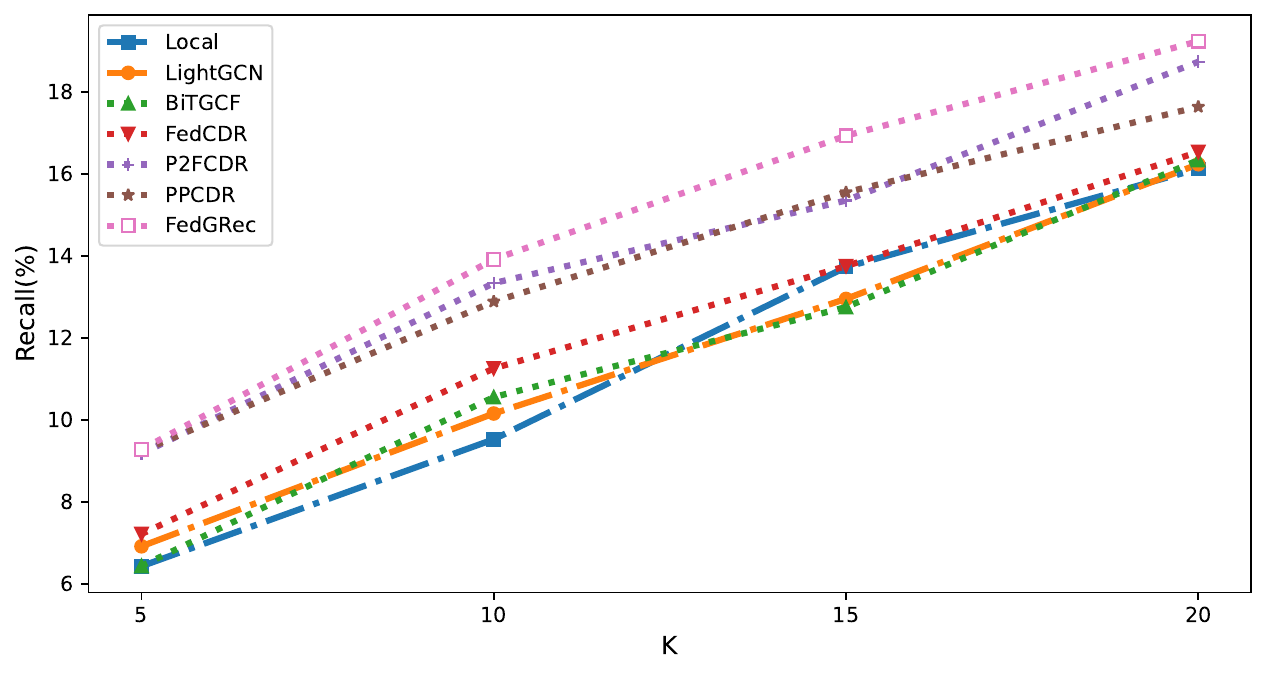}
        \caption{Amazon-Book}
        \label{fig:sub2}
    \end{subfigure}
    \hfill
    \begin{subfigure}[b]{0.3\textwidth}
        \includegraphics[width=\textwidth]{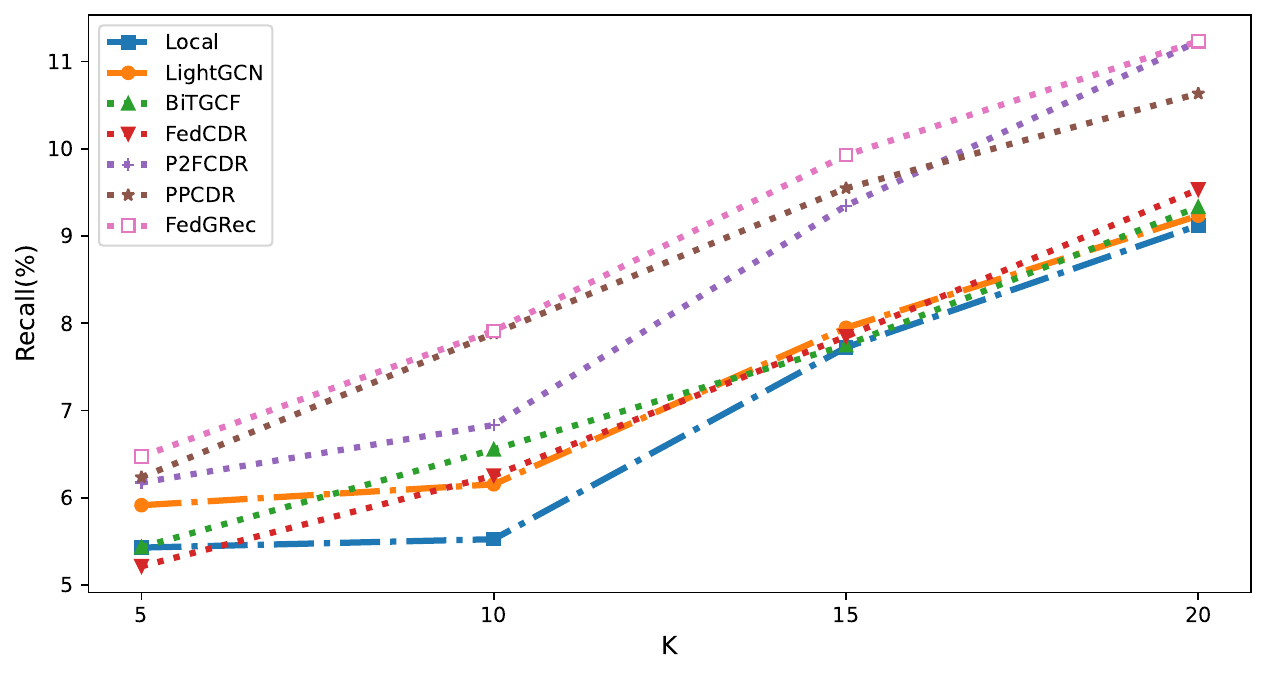}
        \caption{Yelp2018}
        \label{fig:sub3}
    \end{subfigure}
    
    \caption{Performance evaluation with varying numbers of recommended items.}
    \label{fig:Parameter}
\end{figure}

We first evaluate the impact of the number of recommended items ($K$) on model performance across three datasets—Gowalla, Amazon-Book, and Yelp 2018—with detailed results presented in Figure \ref{fig:Parameter}. Our analysis reveals that as $K$ increases, all models exhibit a consistent performance improvement. Notably, the FedGRec model demonstrates superior performance across all tested $K$ values, outperforming the baseline approaches.

% \begin{figure}[!tbp]
% \centering
% \includegraphics[width=0.95\columnwidth]{Figures/Fig8.png}
%     \caption{Performance evaluation with varying numbers of recommended items.}
%     \label{fig:Parameter}
% \end{figure}

\section{Conclusion}\label{}

In this paper, we propose FedGRec, a novel federated graph learning approach that leverages collaborative information from local subgraphs closely associated with users or items to enrich and enhance their representation capabilities. This framework not only effectively models personalized local preferences for given users but also captures global preferences spanning multiple domains. Specifically, FedGRec incorporates a spatiotemporal module that integrates global and local user preferences during private updates within each domain, enabling deep information fusion. Furthermore, we implement a personalized aggregation strategy applied during the federated updating process across multiple domains to preserve global preferences. Experimental results demonstrate that FedGRec achieves superior performance compared to various single-domain and cross-domain baseline methods while strictly maintaining user privacy.

Although FedGRec effectively addresses privacy-preserving challenges in cross-border recommendation systems, several critical issues remain for future work. First, we will investigate more complex cross-border recommendation scenarios where certain users are restricted to interacting within specific domain subsets rather than across the entire user population. Second, to further enhance the performance of privacy-preserving cross-border recommendations, we plan to incorporate more diverse information types, such as text, images, and videos, aiming to achieve significant performance improvements.


\begin{thebibliography}{00}

\bibitem{ref1}
Stevani W, Sudirman L. \textit{Urgensi Perlindungan Data Pengguna Financial Technology terhadap Aksi Kejahatan Online di Indonesia}. Journal of Judicial Review, 2021, 23(2): 197-216.

\bibitem{ref2}
Schreyer M, Sattarov T, Borth D. \textit{Federated and privacy-preserving learning of accounting data in financial statement audits}. in Proceedings of the Third ACM International Conference on AI in Finance. 2022: 105-113.


\bibitem{ref3}
Liao X, Liu W, Zheng X, et al. \textit{Ppgencdr: A stable and robust framework for privacy-preserving cross-domain recommendation}. in Proceedings of the AAAI Conference on Artificial Intelligence. 2023, 37(4): 4453-4461.

\bibitem{ref4}
Bakare S S, Adeniyi A O, Akpuokwe C U, et al. \textit{Data privacy laws and compliance: a comparative review of the EU GDPR and USA regulations}. Computer Science \& IT Research Journal, 2024, 5(3): 528-543.

\bibitem{ref5}
Dorfleitner G, Hornuf L, Kreppmeier J. \textit{Promise not fulfilled: FinTech, data privacy, and the GDPR}. Electronic Markets, 2023, 33(1): 33.

\bibitem{ref6}
Christian K. \textit{GDPR and CCPA: A Comparative Analysis of Their Influence on Data Security and Organizational Compliance}. International Journal of Advanced Engineering Technologies and Innovations, 2024, 6(2): 82-92.

\bibitem{ref7}
Wang L, Sang L, Zhang Q, et al. \textit{A Privacy-Preserving Framework with Multi-Modal Data for Cross-Domain Recommendation}. arXiv preprint arXiv:2403.03600, 2024.


\bibitem{ref8}
C. Zhao, C. Li, R. Xiao, H. Deng, A. Sun, \textit{Catn: Cross-domain recommendation for cold-start users via aspect transfer network}, in: Proceedings of the 43rd International ACM SIGIR Conference on Research and Development in Information Retrieval, 2020, pp. 229–238. 

\bibitem{ref9}
W. Fu, Z. Peng, S. Wang, Y. Xu, J. Li, \textit{Deeply fusing reviews and contents for cold start users in cross-domain recommendation systems}, in: Proceedings of the AAAI Conference on Artificial Intelligence, Vol. 33, 2019, pp. 94–101.

\bibitem{ref10}
Bin Li, Qiang Yang, and Xiangyang Xue. \textit{Can movies and books collaborate? cross-domain collaborative filtering for sparsity reduction}. In Twenty-First International Joint Conference on Artificial Intelligence, 2009. 

\bibitem{ref11}
Weiming Liu, Xiaolin Zheng, Mengling Hu, and Chaochao Chen. \textit{Collaborative filtering with attribution alignment for review-based non-overlapped cross domain recommendation}. In Proceedings of the ACM Web Conference 2022, pages 1181–1190, 2022. 

\bibitem{ref12}
Xiaoyun Zhao, Ning Yang, and Philip S Yu. \textit{Multi-sparse-domain collaborative recommendation via enhanced comprehensive aspect preference learning}. In Proceedings of the Fifteenth ACM International Conference on Web Search and Data Mining, pages 1452–1460, 2022. 

\bibitem{ref13}
Weiming Liu, Chaochao Chen, Xinting Liao, Mengling Hu, Yanchao Tan, Fan Wang, Xiaolin Zheng, and Yew Soon Ong. \textit{Learning accurate and bidirectional transformation via dynamic embedding transportation for cross-domain recommendation}. In Proceedings of the AAAI Conference on Artificial Intelligence, number 8, pages 8815–8823, 2024.

\bibitem{ref14}
J. Cao, X. Lin, X. Cong, J. Ya, T. Liu, B. Wang, \textit{Disencdr: Learning disentangled representations for cross-domain recommendation}. in Proceedings of the 45th International ACM SIGIR Conference on Research and Development in Information Retrieval, 2022, pp. 267–277.

\bibitem{ref15}
J. Zhu, Y. Wang, F. Zhu, Z. Sun, \textit{Domain disentanglement with interpolative data augmentation for dual-target cross-domain recommendation}. in: Proceedings of the 17th ACM Conference on Recommender Systems, 2023, pp. 515–527.

\bibitem{ref16}
Tong Man, Huawei Shen, Xiaolong Jin, and Xueqi Cheng. \textit{Cross-domain recommendation: An embedding and mapping approach}. In IJCAI, volume 17, pages 2464–2470, 2017.

\bibitem{ref17}
Ali Mamdouh Elkahky, Yang Song, and Xiaodong He. \textit{A multi-view deep learning approach for cross domain user modeling in recommendation systems}. In Proceedings of the 24th International Conference on World Wide Web, pages 278–288, 2015. 

\bibitem{ref18}
X. Guo, S. Li, N. Guo, J. Cao, X. Liu, Q. Ma, R. Gan, Y. Zhao, \textit{Disentangled representations learning for multi-target cross-domain recommendation}, ACM Transactions on Information Systems 41 (4) (2023) 1–27. 

\bibitem{ref19}
T. Mukande, \textit{Heterogeneous graph representation learning for multitarget cross-domain recommendation}, in: Proceedings of the 16th ACM Conference on Recommender Systems, 2022, pp. 730–734. 

\bibitem{ref20}
Gong J, Wan Y, Liu Y, et al. \textit{Reinforced moocs concept recommendation in heterogeneous information networks}. ACM Transactions on the Web, 2023, 17(3): 1-27.

\bibitem{ref21}
Cai D, Qian S, Fang Q, et al. \textit{User cold-start recommendation via inductive heterogeneous graph neural network}. ACM Transactions on Information Systems, 2023, 41(3): 1-27.

\bibitem{ref22}
W. Yu, X. Lin, J. Ge, W. Ou, Z. Qin, \textit{Semi-supervised collaborative filtering by text-enhanced domain adaptation}, in Proceedings of the 26th ACM SIGKDD International Conference on Knowledge Discovery \& Data Mining, 2020, pp. 2136–2144.

\bibitem{ref23}
T. Man, H. Shen, X. Jin, X. Cheng, \textit{Cross-domain recommendation: An embedding and mapping approach}, in IJCAI, Vol. 17, 2017, pp. 2464–2470. 

\bibitem{ref24}
P. Li, A. Tuzhilin, \textit{Ddtcdr: Deep dual transfer cross domain recommendation}, in Proceedings of the 13th International Conference on Web Search and Data Mining, 2020, pp. 331–339.

\bibitem{ref25}
Q. Cui, T. Wei, Y. Zhang, Q. Zhang, \textit{Herograph: A heterogeneous graph framework for multi-target cross-domain recommendation}, in ORSUM@ RecSys, 2020. 

\bibitem{ref26}
C. Zhao, C. Li, C. Fu, \textit{Cross-domain recommendation via preference propagation graphnet}, in Proceedings of the 28th ACM international conference on information and knowledge management, 2019, pp. 2165–2168.

\bibitem{ref27}
Heitmann B, Kim J G, Passant A, et al. \textit{An architecture for privacy-enabled user profile portability on the web of data[}, in Proceedings of the 1st International workshop on Information Heterogeneity and Fusion in Recommender Systems. 2010: 16-23.

\bibitem{ref28}
Hecht F V, Bocek T, Bär N, et al. \textit{Radiommender: P2p on-line radio with a distributed recommender system}, in 2012 IEEE 12th International Conference on Peer-to-Peer Computing (P2P). IEEE, 2012: 73-74.

\bibitem{ref29}
Agrawal R, Srikant R. \textit{Privacy-preserving data mining}, in Proceedings of the 2000 ACM SIGMOD international conference on Management of data. 2000: 439-450.

\bibitem{ref30}
McSherry F, Mironov I. \textit{Differentially private recommender systems: Building privacy into the netflix prize contenders}, in Proceedings of the 15th ACM SIGKDD international conference on Knowledge discovery and data mining. 2009: 627-636.

\bibitem{ref31}
Berlioz A, Friedman A, Kaafar M A, et al. \textit{Applying differential privacy to matrix factorization}, in Proceedings of the 9th ACM Conference on Recommender Systems. 2015: 

\bibitem{ref32}
Canny J. \textit{Collaborative filtering with privacy}, in Proceedings 2002 IEEE symposium on security and privacy. IEEE, 2002: 45-57.

\bibitem{ref33}
Di Chai, Leye Wang, Kai Chen, and Qiang Yang. \textit{Secure federated matrix factorization}. IEEE Intelligent Systems, 36(5):11–20, 2020. 

\bibitem{ref34}
Chuhan Wu, Fangzhao Wu, Yang Cao, Yongfeng Huang, and Xing Xie. \textit{Fedgnn: Federated graph neural network for privacy-preserving recommendation}. arXiv preprint arXiv:2102.04925, 2021. 

\bibitem{ref35}
Weiming Liu, Chaochao Chen, Xinting Liao, Mengling Hu, Jianwei Yin, Yanchao Tan, and Longfei Zheng. \textit{Federated probabilistic preference distribution modelling with compactness co-clustering for privacy-preserving multi-domain recommendation}. In Proceedings of the 32rd International Joint Conference on Artificial Intelligence, pages 2206–2214, 2023.

\bibitem{ref36}
Zhiwei Liu, Liangwei Yang, Ziwei Fan, Hao Peng, and Philip S Yu. \textit{Federated social recommendation with graph neural network}. ACM Transactions on Intelligent Systems and Technology (TIST), 13(4):1–24, 2022. 

\bibitem{ref37}
Weiming Liu, Xiaolin Zheng, Chaochao Chen, Mengling Hu, Xinting Liao, Fan Wang, Yanchao Tan, Dan Meng, and Jun Wang. \textit{Differentially private sparse mapping for privacy-preserving cross domain recommendation}. In Proceedings of the 31st ACM International Conference on Multimedia, pages 6243–6252, 2023.

\bibitem{ref38}
Peihua Mai and Yan Pang. \textit{Vertical federated graph neural network for recommender system}. In International Conference on Machine Learning, pages 23516–23535. PMLR, 2023. 

\bibitem{ref39}
Gaode Chen, Xinghua Zhang, Yijun Su, Yantong Lai, Ji Xiang, Junbo Zhang, and Yu Zheng. \textit{Win-win: a privacy-preserving federated framework for dual-target cross-domain recommendation}. In Proceedings of the AAAI Conference on Artificial Intelligence, volume 37, pages 4149–4156, 2023.

\bibitem{ref40}
Shuchang Liu, Shuyuan Xu, Wenhui Yu, Zuohui Fu, Yongfeng Zhang, and Amelie Marian. \textit{Fedct: Federated collaborative transfer for recommendation}. In Proceedings of the 44th International ACM SIGIR Conference on Research and Development in Information Retrieval, pages 716–725, 2021. 

\bibitem{ref41}
Wu Meihan, Li Li, Chang Tao, Eric Rigall, Wang Xiaodong, and Xu Cheng-Zhong. \textit{Fedcdr: federated cross-domain recommendation for privacy-preserving rating prediction}. In Proceedings of the 31st ACM International Conference on Information \& Knowledge Management, pages 2179–2188, 2022. 

\bibitem{ref42}
McPherson M, Smith-Lovin L, Cook J M. \textit{Birds of a feather: Homophily in social networks}. Annual review of sociology, 2001, 27(1): 415-444.

\bibitem{ref43}
Yuan K, Liu G, Wu J, et al. \textit{Semantic and structural view fusion modeling for social recommendation}. IEEE Transactions on Knowledge and Data Engineering, 2022, 35(11): 11872-11884.

\bibitem{ref44}
Chu Y, Chang X, Jia K, et al. \textit{Dynamic sequential graph learning for click-through rate prediction}. arXiv preprint arXiv:2109.12541, 2021.

\bibitem{ref45}
Vaswani A, Shazeer N, Parmar N, et al. \textit{Attention is all you need}. Advances in neural information processing systems, 2017, 30.

\bibitem{ref46}
Feng J, Chen Y, Li F, et al. \textit{How powerful are k-hop message passing graph neural networks}. Advances in Neural Information Processing Systems, 2022, 35: 4776-4790.

\bibitem{ref47}
Chen C, Xu Z, Hu W, et al. \textit{FedGL: Federated graph learning framework with global self-supervision}. Information Sciences, 2024, 657: 119976.

\bibitem{ref48}
Hu P, Lin Z, Pan W, et al. \textit{Privacy-preserving graph convolution network for federated item recommendation}. Artificial Intelligence, 2023, 324: 103996.

\bibitem{ref49}
He X, Deng K, Wang X, et al. \textit{Lightgcn: Simplifying and powering graph convolution network for recommendation}, In Proceedings of the 43rd International ACM SIGIR conference on research and development in Information Retrieval. 2020: 639-648.

\bibitem{ref50}
Liu M, Li J, Li G, et al. \textit{Cross domain recommendation via bi-directional transfer graph collaborative filtering networks}, In Proceedings of the 29th ACM international conference on information \& knowledge management. 2020: 885-894.

\bibitem{ref51}
Meihan W, Li L, Tao C, et al. \textit{Fedcdr: federated cross-domain recommendation for privacy-preserving rating prediction}, In Proceedings of the 31st ACM International Conference on Information \& Knowledge Management. 2022: 2179-2188.

\bibitem{ref52}
Chen G, Zhang X, Su Y, et al. \textit{Win-win: a privacy-preserving federated framework for dual-target cross-domain recommendation}, In Proceedings of the AAAI Conference on Artificial Intelligence. 2023, 37(4): 4149-4156.

\bibitem{ref53}
Tian C, Xie Y, Chen X, et al. \textit{Privacy-preserving cross-domain recommendation with federated graph learning}. ACM Transactions on Information Systems, 2024, 42(5): 1-29.

\bibitem{ref54}
Kingma D P, Ba J. \textit{Adam: A method for stochastic optimization}. arXiv preprint arXiv:1412.6980, 2014.


\end{thebibliography}
\end{document}